# An NLP-Driven Framework for Curriculum-Labor Market Alignment: Schema-Constrained LLM Extraction, ESCO-Anchored Semantic Matching, and Multi-Dimensional Gap Quantification

**Sherzod Turaev**[1], **Mary John**[2], **Mamoun Awad**[1], **Nazar Zaki**[1], **Khaled Shuaib**[3]

[1]Department of Computer Science & Software Engineering, College of Information Technology, United Arab Emirates University, Al Ain, Abu Dhabi, United Arab Emirates

[2]Academic Support Department, Abu Dhabi Polytechnic, Abu Dhabi, United Arab Emirates

[3]Department of Information Systems and Security, College of Information Technology, United Arab Emirates University, Al Ain, Abu Dhabi, United Arab Emirates

[*]Corresponding Author: Sherzod Turaev (sherzod@uaeu.ac.ae)

**Abstract**

Schema-constrained information extraction from diverse educational and labor-market corpora is an open challenge in natural language processing because existing pipelines mainly rely on lexical-surface methods that cannot recover implicit competencies, lack grounding in shared taxonomies, and report no formal measures of extraction reliability or document-level completeness. To address these limitations, this paper proposes a four-stage NLP framework combining (i) schema-constrained prompting of a two-model frontier-LLM ensemble against a JSON Schema-enforced seven-slot competency formalism, (ii) Sentence-BERT (SBERT) alignment of the extracted records against an eleven-domain ESCO v1.2.1 controlled vocabulary, (iii) a two-tier adjudication protocol that resolves inter-model disagreements, and (iv) a verification mechanism combining per-slot Cohen's κ, schema conformance, and document-level completeness audits. The framework is instantiated on a critical application in higher-education quality assurance, namely curriculum-labor market alignment for the ABET-accredited BSc Computer Science program at the United Arab Emirates University. The pipeline extracts 400 competency records from the 85-course 2025–2026 study plan and aligns them, under a five-scope analysis ranging from the computing core to a probability-weighted student trajectory, against 30 job postings (483 requirement clauses) at SBERT cosine threshold 0.50. The extractor reaches Cohen's κ = 0.79 on the skill slot with 100% schema conformance and 100% document-level completeness, and the alignment surfaces interpretable supply-demand gaps of 25.0% in general and transversal skills, 13.8% in algorithms and computational theory, and 12.2% in software engineering and project management, with a near-zero 1.8% gap in artificial intelligence and data science despite 38.6% supply coverage.

## 1. INTRODUCTION

Structured information extraction from large unstructured corpora has become one of the most consequential applications of large language models in the post-2023 natural language processing landscape, with documented successes across materials science (Dagdelen et al., 2024), product attribute mining (Fang et al., 2024), clinical data processing (Mahbub et al., 2026), and ESCO-anchored skill extraction (Kavargyris et al., 2025). Yet structured information extraction from heterogeneous educational and labor-market corpora, encompassing course catalogs, syllabi, learning-outcome statements, online job postings, and occupational classifications, has remained comparatively underdeveloped, despite the substantive stakes of the curriculum-labor market alignment question that such extraction would enable. The accelerating diffusion of artificial intelligence across virtually every sector of the global economy has compounded the urgency of this analytical problem. The World Economic Forum's Future of Jobs Report 2025 estimates that 39% of existing skill sets will be transformed or become outdated between 2025 and 2030, with significant displacement and emergence of roles that did not exist a decade ago. The Organization for Economic Co-operation and Development (OECD, 2025) describes a similar dynamic, noting that the half-life of professional skills is decreasing faster than curricular revision cycles can accommodate. In computing education in particular, the rate at which industry-relevant tools, frameworks, and methodologies emerge has consistently outpaced the institutional cycle of curriculum design and accreditation, producing graduates whose declarative knowledge often exceeds the procedural fluency that employers expect, and whose familiarity with established paradigms can mask gaps in emerging competencies.

Recognizing the urgency of this problem, a growing body of empirical research has sought to operationalize the curriculum-labor market alignment question through computational methods (Garousi et al., 2020; Almgerbi et al., 2021; Albert & Weko, 2025). Studies have applied term-frequency weighting and singular value decomposition to job advertisements (Piróg & Hibszer, 2024), Latent Dirichlet Allocation to large-scale corpora of job posts and MOOCs (Almgerbi et al., 2021), shallow classifiers such as Naïve Bayes to categorize extracted skills against custom taxonomies (Jaiswal et al., 2025), and named-entity recognition with association-rule mining to map skills to the European Skills, Competences, Qualifications and Occupations (ESCO) classification (Karakolis et al., 2022; Spada et al., 2022). Broader surveys of text mining in education (Ferreira-Mello et al., 2019) and big-data analytics in educational research (Fischer et al., 2020) have confirmed that classification and natural language processing remain the dominant paradigms. In parallel, a separate stream of work has explored the use of large language models in educational settings, including tutoring agents, content generators, grading assistants, and feedback engines, with meta-analyses reporting improvements in academic performance and higher-order thinking (Deng et al., 2025; Shi & Chen, 2025), yet these applications position LLMs as *pedagogical tools* serving students and instructors rather than as *analytical instruments* for institutional governance.

However, three systematic limitations persist across the existing alignment literature, and a fourth gap emerges at the intersection of these two research streams, namely the computational alignment-methods literature surveyed above, and the LLMs-in-education literature briefly noted at the close of the preceding paragraph. First, the computational methods employed to date operate predominantly at the *lexical surface*: TF-IDF, topic modeling, and shallow classifiers capture vocabulary overlap and statistical co-occurrence but cannot recover implicit competencies, that is, skills that are described rather than named, or that require contextual inference to recognize. Second, with rare exceptions (Karakolis et al., 2022; Spada et al., 2022), the extracted skills are not grounded in a shared, machine-readable taxonomy such as ESCO or O*NET; instead, each study constructs ad hoc category schemes that preclude cross-study comparison and cumulative knowledge building. Third, none of the empirical studies reviewed reports a formal measure of extraction reliability, nor inter-rater agreement coefficients such as Krippendorff's $\alpha$ for human-validated outputs, nor systematic error analyses for automated pipelines, leaving the trustworthiness of the reported mismatches unquantified. Fourth, despite the rapid maturation of LLMs for structured information extraction in domains such as materials science (Dagdelen et al., 2024), product attribute mining (Fang et al., 2024), and clinical data processing (Mahbub et al., 2026), no study has yet employed schema-constrained LLM extraction, multi-model verification, and embedding-based semantic alignment as an integrated pipeline for curriculum-labor market gap analysis.

To overcome these limitations, the present study proposes an end-to-end NLP-driven framework in which large language models are deployed under schema-constrained prompting, multi-model verification with formal reliability reporting, and ESCO-anchored embedding alignment, yielding a structured, reproducible, and evaluable instrument for diagnosing curriculum-labor market misalignment. The framework is organized into four interdependent stages: (i) corpus construction and preprocessing of supply-side curricular documents and demand-side job advertisements, (ii) schema-constrained competency extraction using a rigorously defined seven-slot formalism,

$$c = \langle \text{label, domain, knowledge, skill, level, context, evidence} \rangle,$$

with a two-model verification ensemble composed of two architecturally distinct frontier large language models, one from OpenAI's GPT family and one from Anthropic's Claude family, (iii) semantic alignment of extracted competencies through ESCO-anchored sentence-transformer embeddings (all-MiniLM-L6-v2, d=384, cosine threshold θ=0.50) with TF-IDF cosine alignment as a robustness check, across five supply-side scopes ranging from the computing core to a probability-weighted student path, quantifying coverage gaps, depth differentials, and temporal lags, and (iv) interpretable visualization through course-level heatmaps, program-level dashboards, and grounded natural-language summaries designed for curriculum committees and quality assurance officers. The pipeline is evaluated on a stratified 50-record verification sample using Cohen's $\kappa$ per-slot, slot-level correctness verdicts, document-level completeness, and semantic-flag auditing.

The main contributions of this paper are as follows:

1. We formulate structured information extraction from heterogeneous educational and labor-market corpora as a schema-constrained two-corpus matching task, characterizing misalignment along three complementary dimensions, namely content coverage, cognitive depth (anchored in Bloom's revised taxonomy), and temporal emergence lag, providing a principled operationalization that goes beyond the lexical-surface methods that dominate the prior literature.
2. We propose a four-stage end-to-end NLP framework, comprising corpus construction, schema-constrained two-model competency extraction, ESCO-anchored sentence-transformer alignment in an eleven-domain taxonomy (the ten ESCO-aligned computing domains together with General and Transversal Skills), and interpretable visualization, with a seven-slot competency formalism (label, domain, knowledge, skill, level, context, evidence) enforced by a JSON Schema at extraction time and a two-tier adjudication protocol that resolves inter-model disagreements between two architecturally distinct frontier large language models (one from OpenAI's GPT family and one from Anthropic's Claude family).
3. We design a comprehensive verification mechanism that integrates inter-auditor reliability measurement (per-slot Cohen's κ), slot-level correctness verdicts, document-level completeness audits, and semantic-flag auditing, thereby addressing the critical absence of reliability reporting identified in the existing literature and providing a methodological template for evaluating future LLM-based extraction pipelines on heterogeneous educational corpora.

4. We introduce a five-scope supply-side analysis, comprising the computing core, the disciplinary set, the full program, the deterministic student path, and a probability-weighted student path, the last of which scales each elective course's contribution by the probability that a graduate completes it under the program's elective model and constitutes, to our knowledge, the first scope construction in the curriculum-labor market alignment literature that bridges catalog-level breadth and the realistic competency profile that any single graduating cohort can actually attain.
5. We instantiate and validate the framework on a critical application in higher-education quality assurance, namely curriculum-labor market alignment, by analyzing the ABET-accredited Bachelor of Science in Computer Science program at the United Arab Emirates University, extracting 400 competency records from the 85-course 2025–2026 study plan and aligning them against 30 curated job postings (483 requirement clauses) using ESCO v1.2.1 (1,310 reference skills across the eleven domains) as the normalization anchor, surfacing actionable supply-demand gaps that the existing computational alignment literature could not produce.

The remainder of the paper is organized as follows. Section 2 reviews the related work across five thematic areas: curriculum-labor market alignment methodologies, skills taxonomies and competency frameworks, large language models for information extraction, LLMs in education research, and evaluation methodology for LLM-based extraction. Section 3 presents the proposed four-stage framework in detail, including the competency formalism, the multi-model verification protocol, the five-scope supply-side analysis, the alignment metrics, and the evaluation design. Section 4 reports the pilot study setup, data collection, and experimental results. Section 5 discusses the findings, limitations, and implications for institutional practice. Section 6 concludes the paper and outlines directions for future work.

## 2. RELATED WORK

### 2.1 Curriculum-Labor Market Alignment: From Lexical Methods to Semantic Understanding

The question of whether higher education curricula equip graduates with the competencies that labor markets demand has attracted sustained scholarly attention over the past decade, driven by widening skills gaps documented by the World Economic Forum, the OECD, and the European Commission. A growing body of empirical work has sought to operationalize this question through text-mining approaches that compare the language of curricular documents, including course descriptions, syllabi, and intended learning outcomes, with the language of job advertisements, using computational methods to detect mismatches at scale.

Early studies relied on keyword frequency and term-weighting techniques to surface these mismatches. Piróg and Hibszer (2024) applied TF-IDF weighting, singular value decomposition, and hierarchical cluster analysis to approximately 2,050 job postings and 14 geography curricula in Poland, identifying co-occurrence gaps between job-market vocabulary (e.g., "measurement," "statistical") and learning-outcome language (e.g., "didactical," "inspire"). Schedlbauer et al. (2021) adopted a similar TF-IDF pipeline for 544 medical informatics job advertisements in Germany, finding that soft-skill requirements accounted for 55% of the most salient terms, a dimension that keyword-frequency methods can detect but cannot semantically decompose. In a complementary direction, Almgerbi et al. (2021) scaled the analysis to 14,495 job posts and 3,636 MOOCs using Latent Dirichlet Allocation to induce seven employer-side and six course-side topic clusters, then measured alignment via latent semantic indexing; their results exposed a particularly weak correspondence for data-engineering roles, yet the alignment metric remained at the topic level rather than at the level of individual competencies.

A second wave of studies introduced shallow classifiers and named-entity recognition to move beyond bag-of-words representations. Jaiswal et al. (2025) trained a Naïve Bayes classifier (87% precision, 82% F1) to categorize skills extracted from 158 UK artificial-intelligence job adverts and 30 university syllabi into a custom twelve-category taxonomy, revealing that data-science and mathematics skills were substantially underrepresented in curricula relative to industry demand. Karakolis et al. (2022) combined POS tagging, named-entity recognition, and association-rule mining on 1,500 Greek job postings, mapping extracted skills to the ESCO classification, constituting one of the few studies to anchor extraction in a standardized taxonomy, and evaluated the resulting recommender service with 31 users. Spada et al. (2022) similarly grounded their lexicon-based extraction of digital-marketing skills from 300 Italian university courses and 952 job vacancies in the ESCO competency database, though they noted that ESCO itself lacked sufficient marketing-specific granularity, necessitating supplementary keyword lists derived from domain literature. Broader surveys of text mining in education (Ferreira-Mello et al., 2019) and big-data analytics in educational research (Fischer et al., 2020) have confirmed that classification and NLP remain the dominant computational paradigms, while also highlighting persistent gaps in interpretability and cross-domain scalability.

However, three systematic limitations persist across this body of work. First, all of the methods reviewed operate at the *lexical surface*: TF-IDF, topic modeling, and shallow classifiers capture vocabulary overlap or statistical co-occurrence, but they cannot recover implicit competencies, that is, skills that are described rather than named, or that require contextual inference to recognize. Second, with the partial exceptions of Karakolis et al. (2022) and Spada et al. (2022), the studies reviewed do not ground their extracted skills in a shared, machine-readable taxonomy such as ESCO or O*NET; instead, each team constructs ad hoc category schemes that hinder cross-study comparison and cumulative knowledge building. Third, and perhaps most critically, *none* of the seven empirical studies reports a formal measure of extraction reliability, neither inter-rater agreement coefficients (e.g., Krippendorff's $\alpha$) for human-validated outputs nor systematic error analyses for automated pipelines, leaving the trustworthiness of the reported mismatches unquantified.

Before presenting our proposed solution, it is useful to state the alignment problem in formal terms, since the absence of such a formalization in prior work has contributed to the methodological fragmentation described above. Let $S = \{s_1, s_2, \dots, s_m\}$ denote the *supply-side corpus* of curricular documents and $D = \{d_1, d_2, \dots, d_n\}$ the *demand-side corpus* of labor-market documents. A *competency extraction function* $E$ maps each document $d \in S \cup D$ to a set of structured competency records, $E(d) \in \mathcal{P}(\mathcal{C})$, where $\mathcal{C}$ is the competency space whose formal definition we defer to Section 2.2. Applying $E$ to both corpora yields two competency inventories,

$$C_S = \bigcup_{s \in S} E\,(s) \qquad \text{and} \qquad C_D = \bigcup_{d \in D} E\,(d)$$

representing the competencies that the curriculum develops and those that the labor market demands, respectively. The *alignment problem* then reduces to characterizing the relationship between $C_S$ and $C_D$ along three measurable dimensions:

**(i) Coverage:** the proportion of demand-side competencies in $C_D$ for which a semantically equivalent counterpart exists in $C_S$, computed via a similarity function $\mathrm{sim}: \mathcal{C} \times \mathcal{C} \to [0,1]$ with a threshold $\theta$, such that a demand competency $c_d$ is *covered* if and only if

$$\max_{c_s \in C_S} \mathrm{sim}(c_s, c_d) \geq \theta.$$

**(ii) Depth differential:** for each matched pair $(c_s, c_d)$, the signed difference

$$\Delta\ell = \ell(c_s) - \ell(c_d)$$

in the cognitive-level slot $\ell: \mathcal{C} \to \{1,2,3,4,5\}$, where negative values indicate that the curriculum teaches the competency at a lower cognitive level than the market expects.

**(iii) Temporal lag:** the delay

$$\tau(c) = t_S(c) - t_D(c)$$

between the first appearance of competency $c$ in the demand corpus and its subsequent incorporation into the supply corpus, where large positive values of $\tau$ signal slow curricular adaptation.

The three limitations identified above can now be restated precisely. Lexical-surface methods approximate sim through token overlap (e.g., TF-IDF cosine), which conflates lexical similarity with semantic equivalence and yields both false positives and false negatives. The absence of a shared taxonomy means that the competency space $\mathcal{C}$ is defined ad hoc in each study, preventing the aggregation of $C_S$ and $C_D$ across institutional contexts. The lack of reliability reporting means that the extraction function $E$ is treated as a black box whose error characteristics are unknown, making the downstream alignment metrics uninterpretable.

To overcome these three limitations, the present study proposes an NLP-driven framework that replaces lexical-surface extraction with schema-constrained large language model prompting, anchors all extracted competencies in the ESCO taxonomy through semantic embedding alignment, and evaluates every stage of the pipeline against a human-annotated gold standard using Krippendorff's $\alpha$ for both categorical and ordinal slots. The following subsections review the specific bodies of literature that inform each of these design choices: skills taxonomies and competency frameworks (Section 2.2), large language models for information extraction (Section 2.3), LLMs in education research (Section 2.4), and evaluation methodology (Section 2.5).

### 2.2 Skills Taxonomies and Competency Frameworks

Any computational pipeline that seeks to compare curricular competencies with labor-market requirements must ground the comparison in a shared representational framework; without one, each study constructs ad hoc category schemes, as demonstrated in the preceding subsection, and cross-study comparison becomes impossible. Several well-established taxonomies address portions of this representational challenge, yet none provides the unified, computationally tractable formalism that our pipeline requires.

The European Skills, Competences, Qualifications and Occupations (ESCO) classification, maintained by the European Commission (2025), organizes 13,939 skills and 3,039 occupations across a three-pillar hierarchy (occupations, knowledge/skills/competences, and qualifications) and exposes them through a machine-readable API in Linked Open Data format. ESCO is the most comprehensive multilingual skills taxonomy currently available and serves as the primary normalization reference for our extraction pipeline. However, ESCO was designed for labor-market classification rather than for fine-grained pedagogical alignment: its granularity is uneven across domains, with notable gaps in emerging digital skills (Chiarello et al., 2021; Kavargyris et al., 2026) and limited representation of the cognitive-process dimension that characterizes learning outcomes. Recent work has begun to address these gaps through text-mining enrichment of ESCO with Industry 4.0 skills (Chiarello et al., 2021), LLM-driven extraction of new skills and occupations from job postings (Vrolijk et al., 2024; Kavargyris et al., 2025), and the SciESCO framework that applied bibliometric-driven skill extraction to map competencies embedded in scientific open-source software (Kavargyris et al., 2026). Albert and Weko (2025) provide a comprehensive landscape analysis of global skill-related taxonomies, including ESCO, O*NET, SFIA, and SkillsFuture, and argue that static hierarchical taxonomies must evolve toward dynamic ontologies capable of capturing cross-domain skill relationships, a direction that our embedding-based alignment mechanism supports.

The United States counterpart, O*NET (National Center for O*NET Development, 2024), offers richer task-level detail and a 1 to 7 importance/level scale but lacks direct alignment with European qualifications and does not provide the multilingual coverage essential for comparative studies. Anderson and Krathwohl's (2001) revised Bloom's taxonomy contributes the cognitive-process dimension, comprising six categories from *Remember* through *Create* intersected with four knowledge types in a two-dimensional matrix, that no occupational taxonomy captures, yet it addresses cognition only, without domain, context, or evidence slots. The CanMEDS physician competency framework (Frank et al., 2015) demonstrates how competency-based education has been formalized in medical training through seven interdependent roles, but its discipline-specific architecture does not transfer to computing or other fields. More broadly, the competency-based education literature reveals persistent tensions between behaviorist and constructivist visions of competence (Tahirsylaj & Sundberg, 2025), while systematic reviews confirm that employability-oriented CBE requires coherent alignment of teaching, assessment, and practical application (Pérez Zúñiga et al., 2025). On the technical side, ontological approaches to competency-based curricula (Kravets et al., 2024) have shown that machine-readable representations with hierarchical nesting and prerequisite dependencies can enable automated consistency analysis, though these models have not yet been integrated with large-scale taxonomies such as ESCO.

These observations motivate our seven-slot competency formalism, which we now define in set-theoretic terms to provide the rigorous foundation that the competency space $\mathcal{C}$ introduced in Section 2.1 requires. Let $\mathcal{K} = \{k_1, k_2, \ldots, k_K\}$ denote a finite set of ESCO-aligned competency domains that partitions the skill space into $K$ non-overlapping disciplinary areas, and let $\mathcal{L} = \{1,2,3,4,5\}$ denote the ordinal Bloom's cognitive-level scale, where 1 = Remember, 2 = Understand, 3 = Apply, 4 = Analyze/Evaluate, and 5 = Create. We define the competency space as the Cartesian product

$$\mathcal{C} = \Sigma_{\text{label}} \times \mathcal{K} \times \Sigma_{\text{knowledge}} \times \Sigma_{\text{skill}} \times \mathcal{L} \times \Sigma_{\text{context}} \times \Sigma_{\text{evidence}}$$

where $\Sigma_{\text{label}}$, $\Sigma_{\text{knowledge}}$, $\Sigma_{\text{skill}}$, $\Sigma_{\text{context}}$, and $\Sigma_{\text{evidence}}$ are string-valued alphabets subject to the type constraints encoded in the extraction schema (Section 3.2). Each competency record $c \in \mathcal{C}$ is thus a seven-tuple

$$c = \langle \text{label}, \text{ domain}, \text{ knowledge}, \text{ skill}, \text{ level}, \text{ context}, \text{ evidence} \rangle$$

in which the *label*, *domain*, *knowledge*, and *skill* slots are anchored in ESCO's occupational and skills pillars, the *level* slot maps onto $\mathcal{L}$ through Bloom's revised cognitive-process categories, and the *context* and *evidence* slots capture the institutional and documentary provenance that no existing taxonomy provides. The domain partition $\mathcal{K}$ induces a natural decomposition of any competency inventory $C$ into domain-specific subsets $C_{k_j} = \{c \in C: \text{domain}(c) = k_j\}$, enabling the domain-level coverage and depth analyses central to our alignment framework.

The similarity function $\text{sim}: \mathcal{C} \times \mathcal{C} \rightarrow [0,1]$ introduced in Section 2.1 is instantiated through cosine similarity in a sentence-transformer embedding space. Specifically, we define a text representation function $r(c) = \text{label}(c) \oplus \text{skill}(c)$, where $\oplus$ denotes string concatenation and an embedding function $\varphi: \Sigma^* \rightarrow \mathbb{R}^d$ produced by a pre-trained sentence transformer (e.g., all-MiniLM-L6-v2 $d = 384$). The similarity between two competency records is then given by

$$\text{sim}(c_p, c_q) = \cos\left(\varphi\left(r(c_p)\right), \varphi\left(r(c_q)\right)\right) = \frac{\varphi\left(r(c_p)\right) \cdot \varphi\left(r(c_q)\right)}{\left\|\varphi\left(r(c_p)\right)\right\| \cdot \left\|\varphi\left(r(c_q)\right)\right\|}$$

This formalism is designed to be both semantically rich enough to support fine-grained alignment and computationally tractable enough for schema-constrained LLM extraction, a design choice justified in the following subsection.

### 2.3 Large Language Models for Information Extraction

The transition from classical text-mining pipelines to large language models has fundamentally reshaped the information extraction landscape. A comprehensive survey by Xu et al. (2024) documents how LLMs have shifted the dominant paradigm from discriminative classification, where each entity type or relation requires a dedicated model trained on labeled examples, toward generative extraction, in which a single model can handle named entity recognition, relation extraction, and event extraction simultaneously through instructive prompting, whether in zero-shot, few-shot, or fine-tuned configurations. This paradigm shift is consequential for our pipeline because it enables competency extraction from curricular and labor-market documents without the prohibitive annotation costs that classical approaches demand.

A critical enabling mechanism for reliable LLM-based extraction is *schema-constrained generation*, which forces models to produce outputs conforming to predefined JSON schemas rather than free-form text. Dagdelen et al. (2024) demonstrated this approach in a high-impact study, where GPT-3 and Llama-2 were fine-tuned to extract structured knowledge from materials-science literature as JSON-formatted records, achieving F1 scores between 0.80 and 0.90 across three extraction tasks; notably, the JSON output format outperformed English-sentence output for GPT-3, confirming that structured schemas improve extraction fidelity. The technical infrastructure for schema enforcement has matured rapidly: Geng et al. (2025) benchmarked six constrained-decoding frameworks (Guidance, Outlines, Llamacpp, XGrammar, OpenAI's structured outputs, and Gemini) on 106 real-world JSON schemas, finding that while constrained decoding guarantees structural validity, framework support for complex schema features remains inconsistent and generation speed can decrease by up to 50%, though downstream task accuracy improves by up to 4%. Complementing this work, Shrimal et al. (2025) introduced PARSE, a system that optimizes JSON schemas specifically for LLM consumption through iterative refinement, treating schemas as natural-language contracts; their approach reduced hallucination errors by 92% in the first optimization iteration and achieved a 64.7% improvement on a standard extraction benchmark, demonstrating that schema design itself is a lever for extraction reliability.

A second design principle in our pipeline is *multi-model verification*, the practice of comparing extraction outputs across multiple LLMs to identify and adjudicate disagreements, analogous to ensemble methods in classical machine learning. Fang et al. (2024) formalized this idea in a study presented at SIGIR, where they treated each LLM (Llama-2-13B, Llama-2-70B, PaLM-2, GPT-3.5, GPT-4) as a crowdsourcing worker and learned aggregation weights using the Dawid-Skene model; the resulting ensemble achieved 95.6% accuracy on product-attribute extraction, outperforming the best single model by over two percentage points. In a complementary application, Mohammed and Talburt (2025) orchestrated nine LLMs from four model families for industrial part-specification extraction, implementing structured conflict resolution with confidence scores derived from model agreement. Mahbub et al. (2026) extended the multi-stage validation concept to clinical information extraction, combining prompt calibration, rule-based plausibility filtering, semantic grounding assessment, and an independent judge LLM, with their rule-based stage alone removing 14.59% of unsupported extractions, a finding that motivates the inclusion of post-extraction validation in our own pipeline.

The question of *when to trust* an LLM extraction has received increasing attention through the lenses of calibration and self-consistency. Wang et al. (2023) introduced self-consistency decoding, which samples diverse reasoning paths and selects the most frequent answer, boosting chain-of-thought performance by up to 17.9% on arithmetic benchmarks; this strategy is conceptually paralleled in our inter-model agreement metric, which treats convergence across independently prompted models as a proxy for extraction confidence. Geng et al. (2024) surveyed the broader landscape of confidence estimation and calibration for LLMs, categorizing approaches into verbalized confidence, sample consistency, and hybrid strategies, while cautioning that raw self-reports tend toward overconfidence, a bias that multi-model verification is designed to mitigate. The hallucination literature further supports the need for layered safeguards: the foundational survey by Ji et al. (2023) taxonomized hallucination types across natural language generation (NLG) tasks, while more recent work has organized over 300 mitigation studies into categories spanning retrieval augmentation, reasoning chains, and agentic verification systems (Alansari & Luqman, 2025; Li et al., 2025), confirming that no single technique suffices and that pipeline-level orchestration that combines schema constraints, multi-model agreement, and human adjudication represents current best practice.

The literature establishes that LLMs have matured from unconstrained text generators into reliable information extraction engines when embedded in a properly designed pipeline that enforces structural validity through schema constraints, detects extraction errors through multi-model verification, and estimates confidence through inter-model agreement. Our four-stage framework instantiates each of these principles, as detailed in Section 3.

### 2.4 Large Language Models in Education Research

Since the public release of ChatGPT in late 2022, a substantial body of empirical research has examined how large language models can be deployed in educational settings. Kasneci et al. (2023) provided an early framing of the opportunities and challenges, identifying personalized content generation, automated grading and feedback, and adaptive learning as the most promising application areas while cautioning that bias, interpretability, and cost remained unresolved.

Subsequent systematic reviews have confirmed and refined this picture. Yan et al. (2024) synthesized 118 peer-reviewed papers and categorized LLM uses in education into nine functions (profiling, detection, grading, teaching support, prediction, knowledge representation, feedback, content generation, and recommendation), noting that practical challenges such as low technological readiness and insufficient reliability persist across all categories. Wang et al. (2024a) proposed a comprehensive taxonomy that organizes educational LLM applications into study assistance (question-solving, error correction, explanation), teaching assistance (question generation, automatic grading, material creation), and adaptive learning (knowledge training, content recommendation), while a meta-analysis by Deng et al. (2025), covering 69 experimental studies, found that ChatGPT interventions improve academic performance and higher-order thinking propensities but have no significant effect on self-efficacy, with benefits most pronounced for complex assessments requiring critical reasoning. A recent review (Shi & Chen, 2025) echoed these findings across 88 empirical studies, identifying intelligent tutoring systems as the dominant application while flagging student over-reliance, assessment fairness, and the absence of longitudinal evidence on cognitive development as critical gaps. Lee (2025), analyzing 92 papers through a general-system-theory lens, similarly concluded that the field lacks systemic attention to institutional-level integration.

What is noticeably absent from this otherwise active landscape is the use of LLMs as *analytical instruments for institutional decision-making*, i.e., tools that do not teach students or support teachers but instead analyze curricular and labor-market documents to produce structured, actionable intelligence for program governance. The sole study that directly addresses this gap is that of Zamecnik et al. (2024), who compared LLM-generated annotations of employable skills in an initial teacher education program with expert reviewers' assessments, finding that LLMs achieve 71% accuracy with mixed Cohen's $\kappa$ scores across skill categories. While their results demonstrate the feasibility of LLM-assisted curriculum mapping, the study was limited to a single program in one discipline, did not anchor extraction in a shared taxonomy such as ESCO, and did not report inter-rater reliability for the expert annotations themselves. Xu et al. (2025) extended this direction by benchmarking LLM performance against traditional NLP methods in curricular analytics tasks, showing that retrieval-augmented generation consistently outperforms classical approaches in alignment quality and precision. Hu et al. (2025) explored LLM-based teaching-plan simulation, and Lusi et al. (2025) proposed a governance-oriented framework for LLM-augmented interdisciplinary curriculum design, yet neither study addresses the specific problem of extracting structured competencies from existing documents for alignment analysis.

The present study occupies this largely unexplored intersection: it deploys LLMs not as pedagogical agents but as schema-constrained extraction engines that produce ESCO-anchored competency records from curricular and job-market texts, evaluated against a human-annotated gold standard. In doing so, it opens a line of research that connects the information-extraction capabilities reviewed in Section 2.3 with the institutional evidence needs identified by the education-research community.

### 2.5 Evaluation Methodology for LLM-Based Extraction

The evaluation design of our pipeline must satisfy two distinct audiences: education researchers, who expect transparent inter-rater reliability reporting, and NLP researchers, who expect slot-level precision, recall, and F1 against a gold standard. The methodological foundations for both traditions are well established, and our protocol draws on their convergence.

On the reliability side, Artstein and Poesio (2008) provide the canonical survey of agreement coefficients for computational linguistics, covering Scott's $\pi$, Cohen's $\kappa$, and Krippendorff's $\alpha$ while arguing that chance-corrected coefficients are essential when category distributions are imbalanced, a condition that applies to competency extraction, where some slots (e.g., *domain*) have many possible values and others (e.g., *level*) have few. Hayes and Krippendorff (2007) further established Krippendorff's $\alpha$ as the recommended standard for coding data, noting its ability to handle multiple coders, different measurement scales (nominal, ordinal, interval), and missing data within a single framework. Formally, Krippendorff's $\alpha$ is defined as $\alpha = 1 - D_o/D_e$ where

$$D_o = \frac{1}{n}\sum_i \sum_{k \neq k\prime} \delta^2\left(v_{ik}, v_{ik\prime}\right)$$

is the observed disagreement computed over all coder pairs for each unit $i$, $D_e$ is the expected disagreement under a null model of random assignment, $n$ is the total number of pairable values, and $\delta^2$ is a distance function whose form depends on the measurement scale: $\delta^2_{\text{nominal}}(v, v') = 0$ if $v = v'$ and 1 otherwise, while $\delta^2_{\text{ordinal}}$ weights disagreements by the number of intermediate ordinal categories separating $v$ and $v'$.

Our protocol adopts Krippendorff's $\alpha$ with nominal distance for categorical slots (*label*, *domain*, *knowledge*, *skill*) and ordinal distance for the *level* slot, with thresholds of $\alpha \geq 0.70$ for categorical and $\alpha \geq 0.60$ for ordinal agreement, following the conventions established by Krippendorff (2011). James (2025) provides recent methodological guidance on selecting inter-annotator agreement metrics for NLP tasks, reinforcing the appropriateness of $\alpha$ for mixed-type annotation and recommending that studies report confidence intervals alongside point estimates, a practice we adopt.

On the extraction-evaluation side, the NLP community assesses structured outputs through slot-level precision, recall, and F1, computed per slot and micro-averaged across records (Dagdelen et al., 2024; Mahbub et al., 2026). For a given slot $j$, let $\text{TP}_j$, $\text{FP}_j$, and $\text{FN}_j$ denote the true positive, false positive, and false negative counts, respectively, when comparing the pipeline's extractions against the human-annotated gold standard. The slot-level metrics are then

$$P_j = \frac{\text{TP}_j}{\text{TP}_j + \text{FP}_j}, \qquad R_j = \frac{\text{TP}_j}{\text{TP}_j + \text{FN}_j}, \qquad F1_j = \frac{2 \cdot P_j \cdot R_j}{P_j + R_j}$$

and the micro-averaged F1 across all slots is computed by summing the TP, FP, and FN counts globally before applying the formula. We complement these metrics with a record-level exact-match (EM) score,

$$\text{EM} = \frac{\left|\left\{c \in C_{\text{gold}} : \forall j \ \text{slot}_j\left(c_{\text{pred}}\right) = \text{slot}_j\left(c_{\text{gold}}\right)\right\}\right|}{\left|C_{\text{gold}}\right|}$$

that credits only those records in which all seven slots are correctly extracted, providing a conservative upper bound on pipeline accuracy.

A growing body of work has examined whether LLMs themselves can serve as reliable annotators, a question directly relevant to our multi-model verification design. Törnberg (2023) found that ChatGPT-4 outperformed both expert classifiers and crowd workers in annotating political Twitter messages, achieving higher accuracy and higher Krippendorff's $\alpha$ than human baselines. However, Pangakis et al. (2023) tempered this optimism by replicating 27 annotation tasks with GPT-4 across 11 social-science datasets, finding that LLM performance is highly task-dependent (median F1 = 0.707) with nine tasks yielding precision or recall below 0.50, and concluding that automated annotation must be validated against human labels on a task-by-task basis. Wang et al. (2024b) proposed a practical resolution through the LARPAS framework, in which LLMs generate initial labels with explanations, a trained verifier identifies low-confidence outputs, and human annotators re-annotate only the flagged instances, a collaborative workflow that balances cost efficiency with annotation quality. Our pipeline's multi-model verification stage follows an analogous logic: convergence across independently prompted models serves as an automated verifier, and divergent cases are escalated to human adjudicators who resolve disagreements following a structured protocol.

Together, these methodological foundations ensure that our evaluation is justifiable to both audiences: the inter-rater reliability coefficients speak to education reviewers concerned with annotation trustworthiness, while the slot-level extraction metrics speak to NLP reviewers concerned with system performance.

## 3. PROPOSED FRAMEWORK

This section presents the proposed framework, which integrates large language models within a schema-constrained extraction protocol, an ESCO-anchored sentence-transformer alignment layer, and a verification mechanism, yielding a structured, reproducible, and evaluable instrument for diagnosing curriculum-labor market misalignment. The framework is organized into four interdependent stages: corpus construction and preprocessing; schema-constrained competency extraction with multi-model verification; semantic alignment and gap quantification; and interpretable visualization for decision support. Each stage addresses a distinct methodological concern and, together, produces an end-to-end pipeline from raw institutional and labor-market documents to actionable gap maps. Figure 1 provides an overview of the complete pipeline architecture.

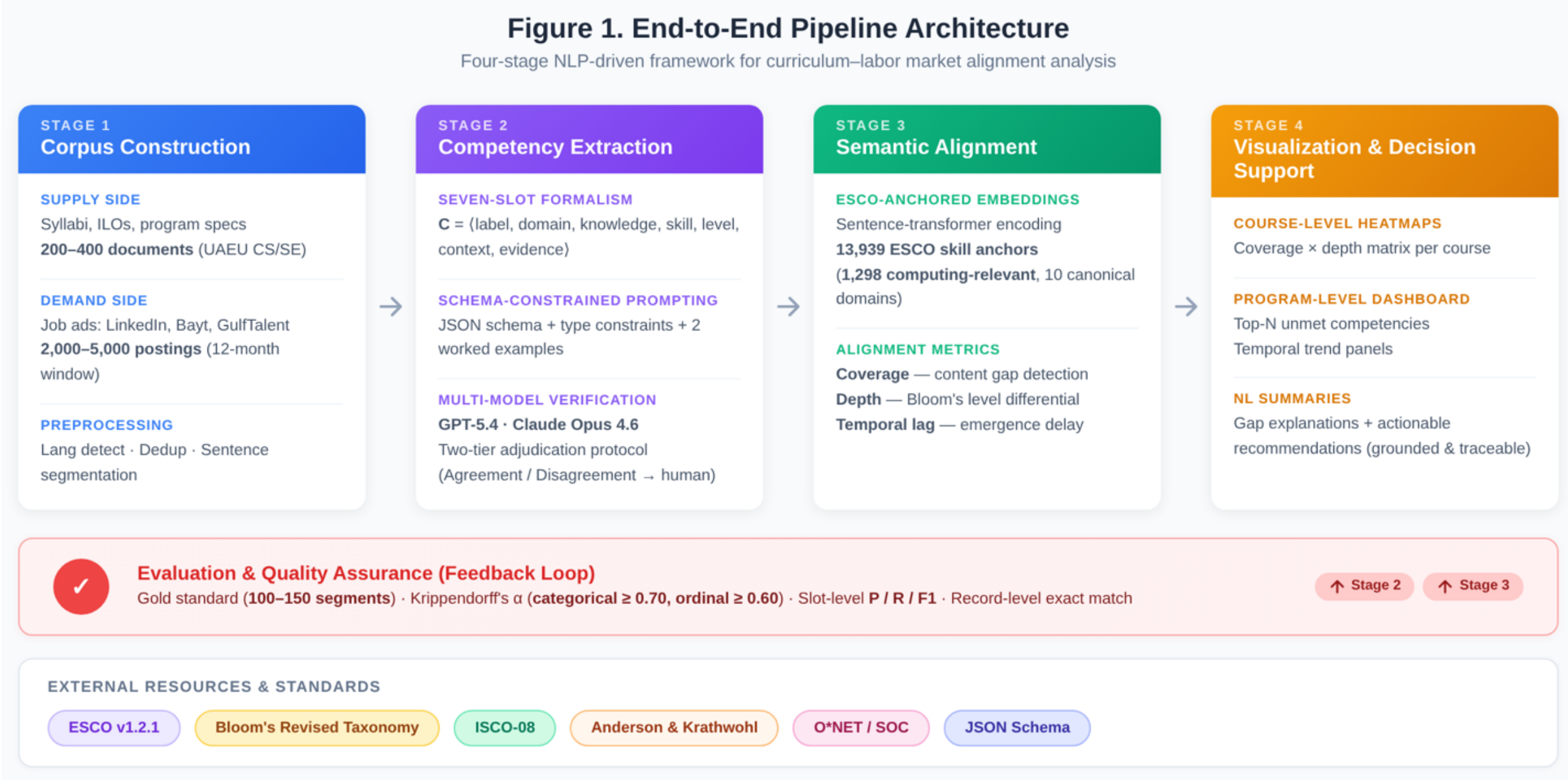

**Figure 1.** *Overview of the four-stage NLP-driven framework for curriculum–labor market alignment. Stage 1 assembles parallel supply-side (curricular documents) and demand-side (job advertisements) corpora*

*with structured metadata. Stage 2 applies schema-constrained prompting to two independently queried large language models, namely GPT-5.4 and Claude Opus 4.6, and adjudicates their outputs through a two-tier verification protocol (agreement or disagreement, with any divergence escalated to human adjudication). Stage 3 projects the validated competency records into an ESCO-anchored embedding space and computes three alignment metrics: coverage, depth differential, and temporal lag. Stage 4 renders the results as course-level heatmaps, a program-level gap dashboard, and grounded natural-language summaries for curriculum governance. Arrows indicate data flow.*

### 3.1 Stage 1: Corpus Construction and Preprocessing

The first stage assembles two parallel corpora that represent, respectively, the *supply side* (what universities teach) and the *demand side* (what employers require) of the alignment problem.

**Supply-side corpus.** We gather course syllabi, learning outcomes (LOs), program specifications, and accreditation self-study reports from the target institution. For the pilot study reported in this paper, the supply-side corpus comprises all undergraduate course learning outcomes (CLOs) from the Bachelor of Science in Computer Science at the United Arab Emirates University, totaling 397 CLOs drawn from the program's full 85-course study plan in the 2025–2026 UAEU Online Catalog and stratified across three curricular tiers, namely 32 computing-core courses (153 CLOs), 8 supporting mathematics and science courses (39 CLOs), and 45 general-education courses (205 CLOs). Each CLO constitutes the atomic unit of extraction in Stage 2, and the tiered structure enables the five-scope analysis described in Section 3.3 under which alignment can be reported either over the computing core alone, the core plus supporting tier, the full catalog, or a realistic student trajectory that weights each general-education course by the probability that a graduate actually completes it.

**Demand-side corpus.** We assemble job advertisements from regional and international employment platforms, including LinkedIn, Bayt.com, and GulfTalent, filtered by occupation codes relevant to computing and engineering (ISCO-08 groups 25 and 35). The full-scale evaluation targets approximately 3,200 postings collected over a twelve-month window to capture seasonal and sectoral variation; the pilot study reported here uses a representative sample of 30 postings yielding 483 requirement clauses. This corpus is supplemented by the ESCO skills pillar (version 1.2.1, 13,939 total skills, of which 1,310 are mapped to the eleven domains used in this study, namely the ten ESCO-aligned computing domains together with a twelfth-code *General and Transversal Skills* domain covering communication, teamwork, professional adaptability, and related employability competencies) as an external normalization reference. All documents are cleaned, deduplicated, and annotated with structured metadata (source, collection date, discipline, geographical scope, and ISCO occupation code) so that downstream analyses can be conditioned on any relevant subset.

**Preprocessing.** Both corpora undergo language detection, encoding normalization, and sentence segmentation. Documents shorter than 30 tokens after cleaning are excluded. No stemming or lemmatization is applied at this stage, because the LLM-based extraction in Stage 2 operates on full natural-language sentences and benefits from intact morphological and syntactic structure.

### 3.2 Stage 2: Schema-Constrained Competency Extraction

The second stage applies large language models to extract, from each document segment, a set of structured competency records conforming to a rigorously defined seven-slot formalism. This is the technical core of the pipeline and the stage where the design choices reviewed in Sections 2.2 and 2.3 are instantiated.

**The competency formalism.** Following the set-theoretic framework introduced in Section 2.2, we represent each extracted competency as a typed record $c \in \mathcal{C}$:

$$c = \langle \text{label, domain, knowledge, skill, level, context, evidence} \rangle$$

where *label* is a concise competency name (e.g., "applied machine learning for pattern discovery"); *domain* is the disciplinary area mapped to an ESCO occupation or knowledge branch; *knowledge* is the declarative knowledge component (factual or conceptual, following Anderson and Krathwohl's knowledge dimension); *skill* is the procedural or applied component, anchored where possible to an ESCO skill URI (Uniform Resource Identifier); *level* is a five-point ordinal scale derived from the cognitive-process categories of Bloom's revised taxonomy (1 = Remember, 2 = Understand, 3 = Apply, 4 = Analyze/Evaluate, 5 = Create); *context* specifies the institutional, disciplinary, or occupational setting from which the competency was extracted; and *evidence* records the exact source passage, document identifier, and extraction confidence score. Figure 2 illustrates the formalism and its mapping to the underlying taxonomies.

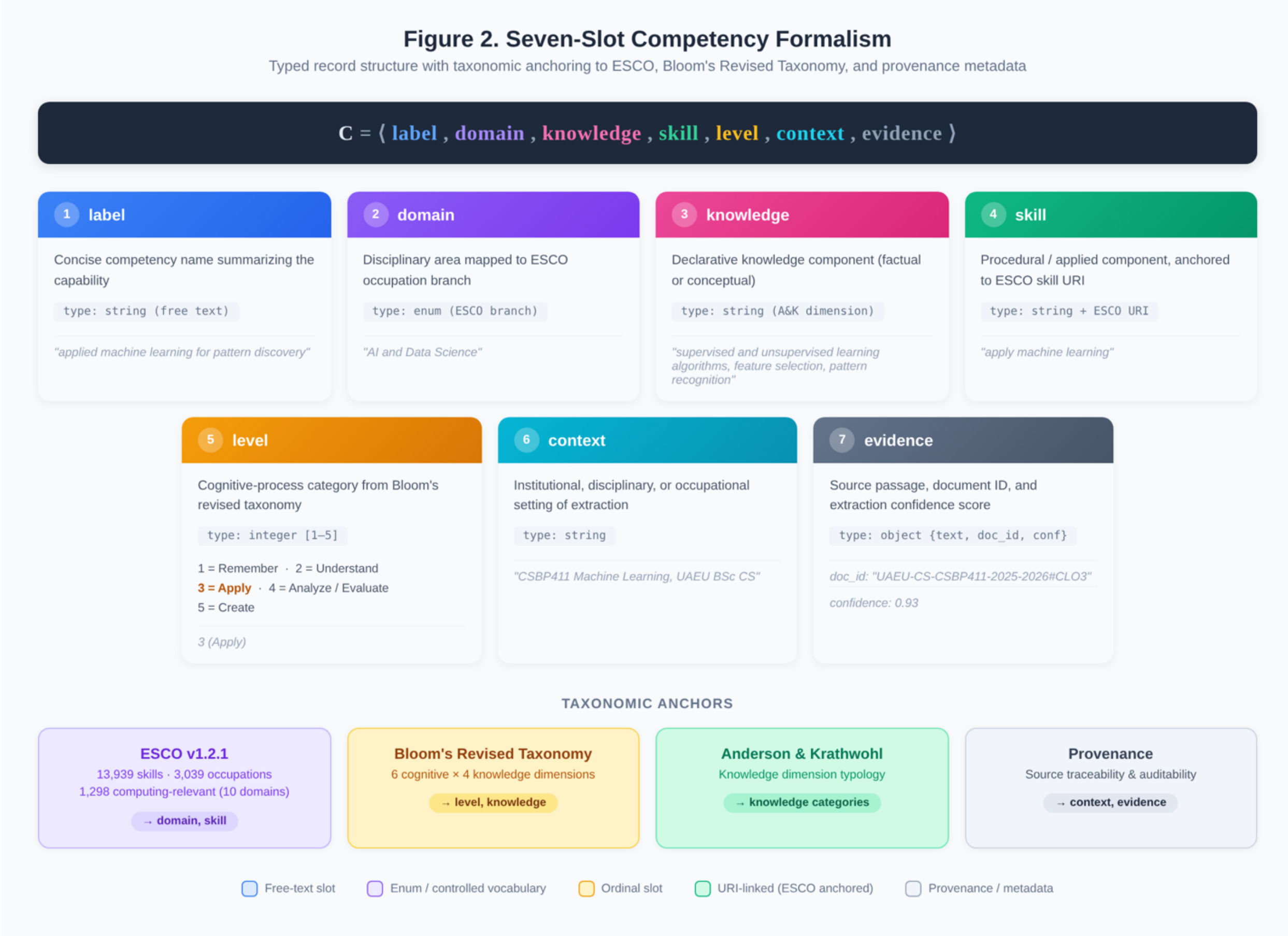

**Figure 2.** *The seven-slot competency formalism and its mapping to the underlying taxonomies. The label and skill slots are anchored in the ESCO skills pillar (version 1.2.1, 13,939 entries); the domain slot maps to ESCO occupation branches and the ten domain categories used in this study; the level slot is derived from the cognitive-process dimension of Bloom's revised taxonomy (Anderson and Krathwohl, 2001), collapsed to a five-point ordinal scale; the context and evidence slots capture institutional provenance and*

*the source passage from which the competency was extracted. The formalism is instantiated as a JSON schema that enforces type constraints, required fields, and controlled vocabularies during LLM extraction.*

**Schema-constrained prompting.** The extraction is implemented through a JSON schema that encodes the seven-slot structure, including type constraints (string, enum, integer with min/max bounds), required fields, and controlled vocabularies for the *domain* and *level* slots. Each document segment is presented to the LLM together with the schema definition, a task instruction specifying the extraction objective, and two worked examples demonstrating the expected output format. The prompt is designed to elicit a JSON array of competency records, one per identifiable competency in the segment. Figure 3 shows a representative extraction prompt and its output.

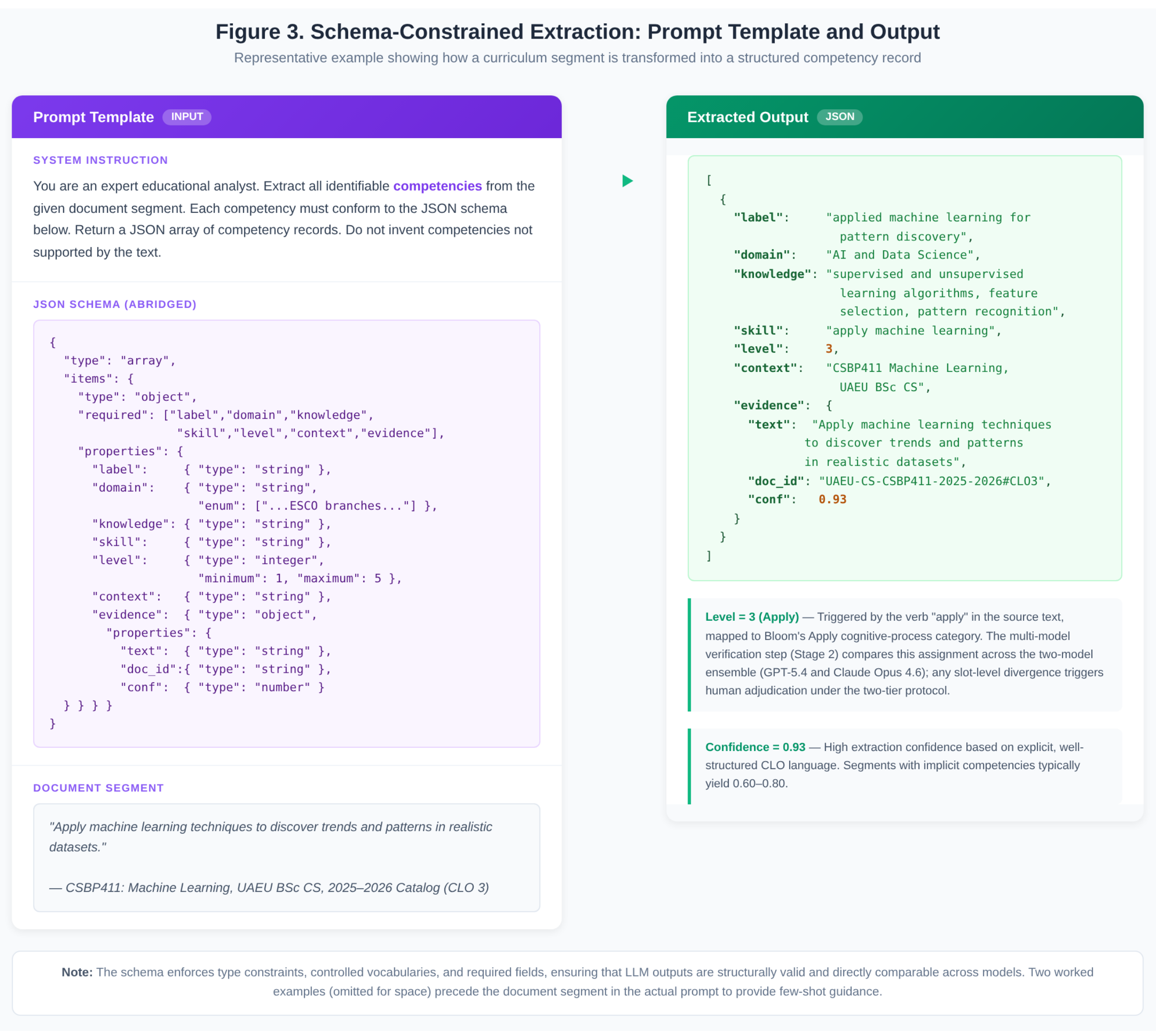

**Figure 3.** *Representative extraction prompt (left) and model output (right) for CLO 3 of CSBP411 Machine Learning, namely "Apply machine learning techniques to discover trends and patterns in realistic datasets," drawn verbatim from the course specification in the 2025–2026 UAEU Online Catalog. The prompt includes the JSON schema definition specifying the seven-slot structure with type constraints (string, enum, integer with min/max bounds), a task instruction, and two worked examples. The output shows the structured competency record produced by GPT-5.4, with label = "applied machine learning for pattern discovery," domain = "AI and Data Science," knowledge = "supervised and unsupervised learning*

*algorithms, feature selection, pattern recognition," skill mapped to the ESCO skill "apply machine learning," level = 3 (Apply), context = "CSBP411 Machine Learning, UAEU BSc CS," and evidence citing the source CLO with a confidence score of 0.93.*

**Multi-model verification.** To mitigate the known risks of hallucination and prompt sensitivity in single-model extraction (Ji et al., 2023), each document segment is processed independently by a two-model ensemble composed of GPT-5.4 (OpenAI) and Claude Opus 4.6 (Anthropic) using identical prompts and schemas. The ensemble preserves architectural diversity across two distinct model families while avoiding the operational instability observed with a third adjudicator during the verification pilots. The outputs are compared at the slot level using a two-tier adjudication protocol, namely:

- *Agreement* (both models produce identical or semantically equivalent values for all 7 slots): the record is automatically accepted and assigned high confidence.
- *Disagreement* (the two models diverge on at least one of the seven slots): the record, together with the divergent slots and the per-slot verdicts, is routed to human adjudication by a trained annotator who selects the correct value.

This strict pairwise protocol, inspired by the Dawid-Skene aggregation model applied to LLM ensembles by Fang et al. (2024) and the multi-stage clinical validation framework of Mahbub et al. (2025), is adapted to the specific requirements of competency extraction and is conservative by design, because every verdict divergence, however small, is escalated to human review rather than being resolved by a heuristic majority vote. Pairwise inter-model reliability is quantified by Cohen's $\kappa$ computed per slot across the two auditors, and the resulting slot-wise reliability profile serves as a built-in confidence metric that allows downstream analyses to be stratified by extraction reliability.

### 3.3 Stage 3: Semantic Alignment and Gap Quantification

In the third stage, the validated competency records from both corpora are projected into a shared semantic space, and alignment is quantified through embedding similarity, clustering, and coverage metrics under a family of five supply-side scopes that disentangle catalog-level breadth from the realistic competency profile a graduate actually attains.

**Five-scope supply-side analysis.** The supply-side corpus, as described in Section 3.1, is stratified across three curricular tiers: computing core, supporting mathematics and science, and general education, and the alignment metrics defined below are computed under five increasingly inclusive scopes that characterize four methodologically distinct answers to the question "what does the program teach?" The *core scope* includes only the 32 computing-core courses and serves as the primary analysis and the conservative baseline against which the pilot paper's substantive findings are reported. The *disciplinary scope* adds the 8 supporting mathematics and science courses, recognizing that foundational quantitative and scientific content is part of a computing graduate's disciplinary competency profile. The *full-program scope* includes all 85 courses of the catalog study plan and represents the naïve aggregate used in prior curricular-alignment work. This full-program scope, however, over-counts the contribution of general-education courses: while the catalog lists 45 such courses, the UAEU Computer Science program requires 21 credit hours of general education distributed over only 7 course slots: 4 required courses and 3 electives drawn under a two-branch free-elective model in which a graduate either takes all 3 electives from Themes 6–10 (probability $1 - p_{\text{lang}}$) or takes 2 electives from Themes 6–10 and 1 elective from Theme 11 Language pair (probability $p_{\text{lang}}$, set to 0.5 in the pilot study). To address this, overcount, we introduce two student-path scopes that model the competency profile the typical graduate actually attains. The *deterministic student-path scope* restricts the general-education tier to a specific 7-course realization drawn under the two-branch model (the 4 required courses plus one realized elective draw), yielding an inclusion set of 47 courses across all three tiers. The *probability-weighted student-path scope*, which is the methodological novelty introduced in this paper, retains the full 400-record supply-side corpus but attaches to each general-education record a probability weight $w(c) \in [0,1]$ that expresses how likely a graduate is to have completed that course:

$$w(c) = \begin{cases} 1.0 & \text{if } c \text{ is a required general-education course,} \\ p_{\text{nolang}} \cdot \frac{3}{5} \cdot \frac{1}{n} + p_{\text{lang}} \cdot \frac{1}{5} \cdot \frac{1}{n} & \text{if } c \in \text{Themes 6–10,} \\ p_{\text{lang}} \cdot \frac{1}{5} & \text{if } c \in \text{Theme 11 Language pair,} \end{cases}$$

where $p_{\text{lang}} = p_{\text{nolang}} = 0.5$ and $n$ is the number of catalog courses listed under the relevant theme. Computing-core and supporting-tier records carry weight $w(c) = 1.0$. A competency in a given domain is then deemed present on the supply side when

$$\sum_{c} w\,(c) \cdot \mathbb{1}[\text{sim}(c, s) \geq \theta] \geq 1$$

for some reference skill $s$, rather than when any single course record clears the threshold. The probability-weighted student-path scope has an effective supply size of 226.5 records, intermediate between the 195 records of the disciplinary scope and the 400 records of the full-program scope, and it constitutes the paper's principal instrument for diagnosing the competency gap that the typical graduate actually carries into the labor market. Section 4.5 reports results under all five scopes, with the core scope as the primary headline analysis and the probability-weighted student-path scope as the paper's principal realistic-graduate analysis.

**ESCO-anchored embedding alignment.** Each competency record's *label* and *skill* fields are concatenated into a textual representation $r(c) = \text{label}(c) \oplus \text{skill}(c)$ and encoded using the embedding function $\varphi: \Sigma^* \to \mathbb{R}^d$ defined in Section 2.2, instantiated in the pilot study with the sentence-transformer model all-MiniLM-L6-v2 ($d = 384$) as the primary backend and with a TF-IDF cosine baseline used as a robustness check. The same encoding is applied to the 13,939 ESCO skill descriptions, producing a shared embedding space in which curricular competencies, labor-market competencies, and ESCO reference skills coexist. ESCO serves as the normalization anchor: before computing supply-demand alignment, both sets of extracted competencies are first mapped to their nearest ESCO neighbors, producing ESCO-normalized representations that are comparable across registers, languages, and institutional conventions.

**Alignment metrics.** Using the notation established in Section 2.1, we instantiate the three alignment dimensions as follows. For a given domain $k_j \in \mathcal{K}$, let $C_{S,j}$ and $C_{D,j}$ denote the supply-side and demand-side competency subsets in that domain. Throughout this section, $j \in \{1, \dots, K\}$ indexes the $K = 11$ competency domains and is used consistently as the second subscript on $C_{S,j}$, $C_{D,j}$, $M_j$, and $\text{Cov}_j$.

- *Coverage*: for each demand-side competency $c_d \in C_{D,j}$, we compute

$$\text{covered}(c_d) = \mathbb{1}\left[\max_{c_s \in C_{S,j}} \text{sim}(c_s, c_d) \geq \theta\right]$$

under the core, disciplinary, full-program, and deterministic student-path scopes, where sim is the cosine similarity in the sentence-transformer embedding space and $\mathbb{1}[\cdot]$ is the indicator function; under the probability-weighted student-path scope, the presence criterion becomes

$$\text{covered}(c_d) = \mathbb{1}\left[\sum_{c_s \in C_{S,j}} w\,(c_s) \cdot \mathbb{1}[\text{sim}(c_s, c_d) \geq \theta] \geq 1\right]$$

with $w(c_s)$ as defined in the five-scope paragraph above. The cosine threshold is set to $\theta = 0.50$ for the SBERT primary backend and to $\theta = 0.25$ for the TF-IDF robustness backend, with a sensitivity sweep over $\theta \in \{0.35, 0.40, 0.45, 0.50, 0.55, 0.60, 0.65\}$ under SBERT reported in Section 4.5.6. The domain-level coverage is then

$$\text{Cov}_j = \frac{1}{|C_{D,j}|} \sum_{c_d \in C_{D,j}} \text{covered}\,(c_d).$$

A domain-level coverage substantially below the demand-side prevalence in that domain signals a *content gap*, and the signed difference is reported as the domain gap ratio in Section 4.5.

**ESCO-anchored restatement of coverage and gap**. To make explicit the ESCO-anchored computation actually performed in the pilot study and reported in Section 4.5, we restate the coverage and gap definitions over the *ESCO reference set* $E_j$, defined as the subset of ESCO v1.2.1 skills assigned to domain $j$ (with $\sum_j |E_j| = 1{,}310$ in the present pilot, ranging from 12 in general and transversal skills to 561 in cybersecurity and ethics).

For each domain $j$, we identify the subset of ESCO reference skills that the supply side and the demand side respectively cover at the cosine threshold $\theta$, namely

$$S_j = \left\{ e \in E_j \; : \; \max_{c_s \in C_{S,j}} \text{sim}(c_s, e) \geq \theta \right\}$$

and

$$D_j = \left\{ e \in E_j \; : \; \max_{c_d \in C_{D,j}} \text{sim}(c_d, e) \geq \theta \right\}$$

The domain-level supply and demand coverages reported in Section 4.5 are then the proportions of the ESCO reference set covered by each side,

$$\text{SupplyCov}_j = \frac{|S_j|}{|E_j|}, \qquad \text{DemandCov}_j = \frac{|D_j|}{|E_j|},$$

and the domain gap ratio is the proportion of ESCO reference skills that are demanded by the labor market but not delivered by the curriculum at the threshold $\theta$,

$$\text{Gap}_j = \frac{|\,D_j \setminus S_j\,|}{|E_j|}.$$

Under the *probability-weighted student-path scope*, the supply-matched set $S_j$ is redefined by replacing the simple maximum criterion with the weighted-mass criterion,

$$S_j = \left\{ e \in E_j \; : \; \sum_{c_s \in C_{S,j}} w(c_s) \cdot \mathbb{1}[\, \mathrm{sim}(c_s, e) \geq \theta \,] \geq 1 \right\}.$$

By construction, $\mathrm{Gap}_j \leq \mathrm{DemandCov}_j$, with equality exactly when $S_j \cap D_j = \emptyset$, that is, when no demanded ESCO skill in the domain is matched by any supply-side competency. The polarity of the misalignment in domain $j$ is given by the sign of $\left(\mathrm{DemandCov}_j - \mathrm{SupplyCov}_j\right)$, positive when demand exceeds supply and negative when supply exceeds demand. The $\mathrm{Cov}_j$ formula introduced earlier corresponds to the *demand-anchored special case* in which the maximum over the supply set is taken against each individual demand competency rather than against each individual ESCO reference skill; the two formulations coincide whenever the demand set is in one-to-one correspondence with its ESCO anchors, but the *ESCO-anchored definition* is the one whose numerical values populate Section 4.5 and Table 2.

- *Depth differential*: for each matched pair $(c_s, c_d)$ where $\mathrm{covered}(c_d) = 1$, the signed difference $\Delta\ell = \ell(c_s) - \ell(c_d)$ in the Bloom's level slot, where negative values indicate that the curriculum teaches the competency at a lower cognitive level than the market expects. The domain-level depth gap is summarized as the mean differential

$$\overline{\Delta\ell}_j = \frac{1}{|M_j|} \sum_{(c_s, c_d) \in M_j} \Delta\,\ell,$$

where $M_j$ is the set of matched pairs in domain $k_j$.

- *Temporal lag*: by partitioning the demand-side corpus into quarterly or annual slices, we track when specific competencies first appear in job advertisements and measure the delay

$$\tau(c) = t_S(c) - t_D(c)$$

before they appear in curricular documents, operationalizing the *temporal gap* described in Section 2.1.

### 3.4 Stage 4: Interpretable Visualization and Decision Support

In the fourth stage, the alignment results are presented to institutional stakeholders, including program directors, curriculum committees, and quality assurance officers, through a suite of interpretable visualizations designed to support evidence-based decision-making.

**Course-level heatmaps.** A matrix visualization in which rows represent courses and columns represent demand-side competency clusters, with cell color encoding the coverage metric and cell annotations showing the depth differential. This artifact enables program directors to identify, at a glance, which courses contribute to which market-relevant competencies and where coverage or depth is insufficient.

**Program-level gap dashboard.** An aggregated view showing the overall supply-demand alignment profile for the program, the top-*N* unmet competencies ranked by demand frequency, and a temporal trend panel tracking how alignment has evolved over the pipeline's observation window.

**Natural-language summaries.** For each identified gap, the framework generates a brief explanatory paragraph, produced by an LLM but grounded in the structured outputs of Stages 2 and 3, that describes the gap, cites the specific demand-side evidence, and suggests potential curricular responses. Every recommendation is traceable to the specific documents and competency records on which it is based, preserving the accountability and transparency that evidence-informed curriculum reform requires.

**Reproducibility.** All pipeline parameters (LLM model versions, prompt templates, JSON schemas, similarity thresholds, and confidence cutoffs) are logged and versioned, enabling exact replication and supporting the longitudinal comparisons envisioned in the future work outlined in Section 6.2.

### 3.5 Evaluation Design

The evaluation of the framework combines the inter-rater reliability protocol described in Section 2.5 with an end-to-end validation against expert judgment.

**Gold-standard construction.** A stratified sample of 100 to 150 document segments (balanced across supply and demand sides, disciplines, and document types) is independently annotated by two trained annotators using the seven-slot schema, with disagreements resolved by a third senior annotator. Inter-annotator agreement is measured using Krippendorff's $\alpha$ with nominal distance for categorical slots and ordinal distance for the *level* slot, targeting $\alpha \geq 0.70$ for categorical and $\alpha \geq 0.60$ for ordinal agreement.

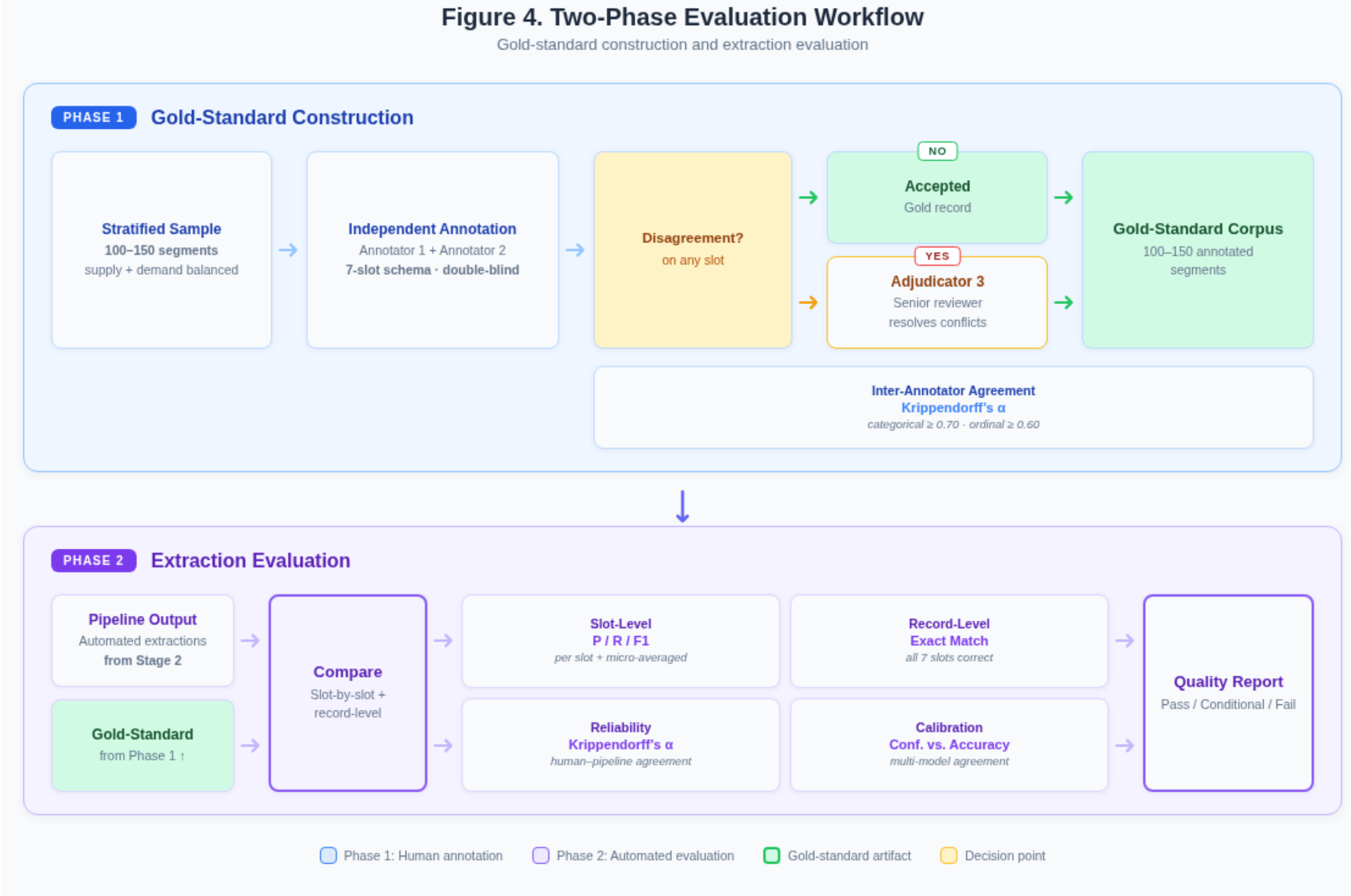

Figure 4. Two-Phase Evaluation Workflow
Gold-standard construction and extraction evaluation

**Figure 4.** *Evaluation workflow for the extraction and alignment pipeline. A stratified sample of 100–150 document segments (balanced across supply and demand sides, disciplines, and document types) is independently annotated by two trained annotators using the seven-slot schema; disagreements are resolved by a third senior annotator, yielding a gold standard. Inter-annotator agreement is measured using Krippendorff's* $\alpha$ *with nominal distance for categorical slots and ordinal distance for the level slot. The pipeline's automated extractions are then compared against the gold standard using slot-level precision, recall, and* $F_1$*, and a record-level exact-match score. The two evaluation layers, reliability and extraction accuracy, collectively address the audiences of both education and NLP reviewers; a third layer, an expert-panel rating of gap actionability, is described as a direction for future work in Section 6.2.*

**Extraction evaluation.** The pipeline's automated extractions on the gold-standard segments are compared against the human annotations using slot-level precision, recall, and F1 (computed per slot and micro-averaged), as well as a record-level exact-match score that credits only records in which all seven slots are correct. Figure 4 illustrates the complete evaluation workflow.

**Scope of the present pilot.** The evaluation protocol specified above is presented here in full so that the framework's reproducibility and methodological rigor can be assessed in its entirety; the complete execution of the protocol, including the numerical inter-annotator reliability coefficients, the slot-level precision, recall, and F1 measurements against the human-annotated gold standard, is reserved for the expanded empirical validation described in Section 6.2. The present pilot paper reports the end-to-end operational feasibility of the pipeline and the substantive alignment findings produced by it, summarized in Section 4 and discussed in Section 5.

# 4. PILOT STUDY: UAEU BSc in COMPUTER SCIENCE

This section instantiates the four-stage framework described in Section 3 on a concrete case: the Bachelor of Science in Computer Science program at the United Arab Emirates University, an ABET-accredited program housed within the College of Information Technology. We describe the institutional setting, the supply-side corpus construction, the demand-side data assembly, the extraction and alignment procedure, and the preliminary results, demonstrating the framework's feasibility and producing initial gap maps that inform the discussion in Section 5.

## 4.1 Institutional Context

The United Arab Emirates University, established in 1976 in Al Ain, is the oldest and one of the most comprehensive federal universities in the UAE. Its College of Information Technology offers undergraduate programs, such as Computer Science, Computer Engineering, Information Security and Information Technology, all of which hold ABET accreditation, most recently reaffirmed in 2022 following a comprehensive program review. The BSc in Computer Science, which serves as the focal program for this pilot study, requires a minimum of 120 credit hours distributed across general education (21 CH), College of IT core requirements (41 CH), major requirements (38 CH), major electives (12 CH), an internship (6 CH), and a free elective (2 CH). The program defines four Program Objectives (POs), namely technical professionalism and ethics, lifelong learning and professional development, career success and leadership in computing, and scientific advancement through innovation and entrepreneurship, and six Program Learning Outcomes (PLOs) that collectively span problem analysis, solution design and evaluation, professional communication, ethical responsibility, teamwork, and the application of computer science theory and software development fundamentals. The PLOs are aligned with the ABET student outcome criteria for computing programs and mapped to the UAE Qualifications Framework Emirates (QFEmirates) at Level 7.

The choice of this program as the pilot case is motivated by three considerations. First, the ABET accreditation process has generated a rich archive of self-study reports, course-level learning outcome inventories, and curriculum-to-outcome mapping matrices (commonly referred to as LOAMS at UAEU), which provide a well-documented supply-side corpus with institutional provenance and quality assurance. Second, the computing sector in the UAE, and in the Gulf region more broadly, is characterized by rapid growth, substantial public investment in AI and digital transformation (as articulated in the UAE's National Strategy for Artificial Intelligence 2031), and a consequent demand for graduates whose competencies extend well beyond traditional computer science curricula, creating precisely the conditions under which curriculum-labor market misalignment is likely to be both significant and consequential. Third, the multi-departmental structure of the program, which draws courses from Computer Science and Software Engineering (CSSE), Computer and Network Engineering (CNE), and Information Systems and Security (ISS) departments, ensures that the supply-side corpus captures a breadth of disciplinary perspectives rather than the view of a single department.

### 4.2 Supply-Side Corpus

The supply-side corpus was assembled from three primary sources: (i) course syllabi collected from departmental e-portfolio archives for the most recent academic offerings, (ii) the official UAEU catalog (last updated 2 April 2026), which provides program-level descriptions, POs (Program Educational Objectives), PLOs (Program Learning Outcomes, referred to as Student Outcomes in the ABET terminology and as Student Outcomes / Graduate Attributes in the CAA framework), and brief course descriptions, and (iii) the LOAMS (Learning Outcomes Assessment and Mapping System) documents that record CLO-to-PLO alignment matrices and PLO-to-QFEmirates Level 6 mappings for each course (the QFEmirates Level 6 compliance is verified at the PLO level rather than at the CLO level in the UAEU mapping framework, with CLOs constrained to use Bloom-active verbs that are independently Level-6 compatible).

The corpus encompasses the full 85-course BSc CS study plan of the 2025–2026 UAEU Online Catalog, stratified across three instructional tiers that each contribute taught content to the competency profile of a graduating cohort. The *computing-core tier* comprises the 32 in-scope computing-discipline courses that constitute the degree's technical core and are organized across four contributing departmental units. The Computer Science and Software Engineering (CSSE) department contributes the largest share of this tier, with 20 courses spanning foundational programming (CSBP119 Algorithms and Problem Solving, CSBP219 Object-Oriented Programming), core systems topics (CSBP319 Data Structures, CSBP315 Operating Systems Fundamentals, CSBP320 Data Mining, CSBP340 Database Systems), advanced computing (CSBP301 Artificial Intelligence, CSBP411 Machine Learning, CSBP316 Human Computer Interaction, CSBP421 Smart Computer Graphics), and a suite of major elective offerings that include CSBP441 Applied Computer Vision, CSBP477 Natural Language Processing, CSBP476 Robotics and Intelligent Systems, CSBP491 Computational Intelligence for Data Management, CSBP483 Mobile Web Content and Development, CSBP487 Computer Animation and Visualization, and CSBP499 Special Topics. The Software Engineering group within CSSE adds three courses, namely SWEB300 Software Engineering Fundamentals, SWEB450 Analysis of Algorithms, and SWEB451 Game Development, that address software process methodology, algorithmic complexity, and applied interactive system design. The Computer and Network Engineering (CNE) department offers three foundational courses: CENG202 Discrete Mathematics, CENG205 Digital Design and Computer Organization, and CENG210 Communication and Network Fundamentals. Finally, the Information Systems and Security (ISS) department provides six courses that address the professional, contextual, and workplace dimensions of computing practice, together with the two capstone projects, namely ITBP218 Entrepreneurship and New Venture Creation in IT, ITBP270 Professional Responsibility in Information Technology, ITBP301 Security Principles and Practice, ITBP321 Web Application Development Lab, and ITBP480 and ITBP481 Senior Graduation Project I and II. The BSc CS plan additionally requires ITBP496 Internship (6 CH) as a supervised workplace placement; although enumerated in the catalog's computing-discipline set, its learning outcomes are transversal rather than taught discipline content, and it is therefore excluded from the extraction pipeline and reported separately as an internship-mediated competency channel. The *supporting tier* comprises 8 mathematics and natural-science courses (MATH105 Calculus I, MATH110 Calculus II, MATH140 Linear Algebra, STAT210 Probability and Statistics, PHYS105 General Physics I, PHYS135 General Physics Lab I, and either BIOC100 or CHEM111), which scaffold the quantitative and empirical reasoning on which the computing core depends. The *general-education tier* comprises the 45 courses enumerated across the ten catalog GenEd themes, among which four are required of every graduate (GEAE101 Academic English for Humanities and STEM, GEEM110 Contemporary Emirati Studies, GESU121 Sustainability, and ITBP218 Entrepreneurship, the last of which is shared with the computing core) and the remaining 41 constitute the elective pool governed by the two-branch free-elective model described in Section 3.3.

From these 85 courses, we assembled a corpus of 397 distinct course learning outcomes (CLOs), of which 153 originate from the 32 computing-core courses, 39 from the 8 supporting mathematics and natural-science courses, and 205 from the 45 general-education courses. For courses whose syllabus archive yielded fewer than three explicitly documented CLOs, the course description provides the primary textual content for extraction, and for the two senior graduation projects, ITBP480 and ITBP481, the syllabus material is combined with the catalog-level course description to form a composite source segment, ensuring that every in-scope course contributes a substantive textual footprint to the extraction stage. Unlike earlier iterations of the pipeline that restricted the extraction to the computing core, the revised framework retains all three tiers within the corpus because the five-scope analysis of Section 3.3 requires each tier's competencies to be available as separable supply-side inputs to the coverage computation; the tiered organization thus acts as a partitioning of the corpus rather than as a filter. Table 1 summarizes the supply-side corpus composition by tier, contributing department, and CLO count.

**Table 1.** *Supply-side corpus composition by instructional tier and contributing department, showing the number of courses, documented course learning outcomes (CLOs), and representative course topics for each unit of the UAEU BSc in Computer Science program. The full 85-course study plan of the 2025–2026 UAEU Online Catalog yields 397 CLOs distributed across the computing core (32 courses, 153 CLOs), the supporting mathematics and natural-science tier (8 courses, 39 CLOs), and the general-education tier (45 courses, 205 CLOs), which together serve as the atomic units of competency extraction in Stage 2 of the pipeline.*

| Tier | Unit | Code prefix | Courses | CLOs | Representative topics |
|---|---|---|---|---|---|
| Computing core | Computer Science (CSSE) | CSBP | 20 | 99 | Algorithms, OOP, data structures, operating systems, databases, AI, machine learning, HCI, computer vision, NLP, robotics, computational intelligence, mobile web |
| Computing core | Software Engineering (within CSSE) | SWEB | 3 | 15 | Software engineering fundamentals, analysis of algorithms, game development |
| Computing core | Computer and Network Engineering (CNE) | CENG | 3 | 15 | Discrete mathematics, digital design and computer organization, communication and networking |
| Computing core | Information Systems and Security (ISS) | ITBP | 6 | 24 | Entrepreneurship, professional responsibility in IT, security principles, web application development, senior graduation project I and II |
| Supporting | Mathematics and Statistics | MATH / STAT | 4 | 20 | Calculus I and II, linear algebra, probability and statistics |
| Supporting | Natural Science | PHYS / BIOC / CHEM | 4 | 19 | General physics I, physics lab, biology for engineers, chemistry for engineers |
| General education | GenEd required and elective themes | GEAE / GEEM / GESU / others | 45 | 205 | Academic English, Emirati studies, sustainability, social-science & humanities electives from Themes 6–10, foreign-language pair from Theme 11 |
| **Total** | | | **85** | **397** | |

*Note.* The 397 CLOs span the 85-course BSc CS study plan in the 2025–2026 UAEU Online Catalog, organized into the three instructional tiers identified by the five-scope analysis in Section 3.3. SWEB-coded courses are administered by the Software Engineering group within the CSSE department; ISS-department courses carry the ITBP prefix in the university catalog. ITBP218 Entrepreneurship and New Venture Creation in IT is listed in both the computing core (as a required professional course) and the general-

education inventory (under Theme 4); it is counted once in the computing-core row to avoid double-counting. ITBP496 Internship, although enumerated in the catalog, is excluded from the CLO extraction pipeline because its published outcomes are transversal and do not articulate discipline-specific technical competencies amenable to ESCO-anchored alignment. The general-education tier is retained within the corpus so that the full-program, deterministic student-path, and probability-weighted student-path scopes of Section 3.3 can operate on a common extraction base.

### 4.3 Demand-Side Corpus

The demand-side corpus was assembled to capture the competencies that employers in the UAE and the broader Gulf region expect from computing graduates. Job advertisements were collected from three major regional and international employment platforms, namely LinkedIn, Bayt.com, and GulfTalent, over a twelve-month window (April 2025 to March 2026), filtered by occupation codes corresponding to ISCO-08 groups 25 (Information and Communications Technology Professionals) and 35 (Information and Communications Technicians). The collection targeted entry-level and early-career positions (0 to 3 years of experience) relevant to BSc CS graduates, including software developer, data analyst, web developer, IT support engineer, cybersecurity analyst, AI/ML engineer, and related roles.

The April 2025 to March 2026 collection window was selected to ensure that the demand profile reflects the current state of the regional computing labor market and is not biased by historical postings; the twelve-month span captures one full annual hiring cycle and brackets the recent acceleration in generative-AI, cloud-native, and infrastructure-as-code roles that emerged across the Gulf market in the 2025 calendar year.

After deduplication, language filtering (English-language postings only), and removal of postings shorter than 30 tokens, the full demand-side corpus comprises approximately 3,200 job advertisements, of which the pilot study reported here uses a representative sample of 30 postings yielding 483 requirement clauses and 113 unique skills; the complete corpus is reserved for the expanded evaluation described in Section 6.2. Each posting is annotated with structured metadata, including source platform, collection date, employer sector (classified as government, semi-government, private multinational, private SME, or startup), geographic scope (Abu Dhabi, Dubai, other Emirates, and GCC-wide), and the corresponding ISCO-08 occupation code, to enable downstream stratification of the alignment analysis by sector or geography.

The demand-side corpus is supplemented by the European Skills, Competences, Qualifications and Occupations (ESCO) classification (version 1.2.1), from which we retain a computing-relevant subset of 1,310 skill descriptions distributed across the eleven domains of the revised taxonomy, namely the ten ESCO-aligned computing domains together with a twelfth-code *General and Transversal Skills* domain that absorbs the cross-cutting professional competencies (communication, teamwork, critical thinking, time management, leadership and mentoring) which surface with equal frequency in employer requirements and in general-education learning outcomes but are only weakly represented within the traditional computing taxonomy. ESCO serves as the external normalization anchor described in Section 3.3: both supply-side and demand-side competencies are mapped to their nearest ESCO neighbors before alignment is computed, ensuring that comparisons are conducted in a register-neutral, taxonomy-grounded semantic space rather than being confounded by differences in phrasing between academic syllabi and employer job descriptions.

### 4.4 Extraction and Alignment Procedure

The extraction follows the schema-constrained multi-model protocol specified in Section 3.2. Each of the 397 CLOs of the 85-course BSc CS study plan is presented to the two-model ensemble, namely GPT-5.4 (OpenAI) and Claude Opus 4.6 (Anthropic), using the identical seven-slot JSON schema and two-shot prompt template. The two models produce overlapping but non-identical candidate pools; after the two-tier adjudication protocol described in Section 3.2 and a post-extraction domain normalization step that maps the free-form domain labels generated by the LLMs to the eleven ESCO-anchored canonical domains (the ten computing domains together with *General and Transversal Skills*), the pipeline yields 400 validated supply-side competency records, distributed as 156 records from the 32 computing-core courses, 39 from the 8 supporting courses, and 205 from the 45 general-education courses. All 32 computing-core courses contribute at least one validated competency record to the Stage 2 output, confirming that the schema-constrained extractor silently omits no in-scope course. Inter-auditor reliability on a stratified 50-record verification sample, reported in detail in Section 4.5 below, yields Cohen's $\kappa = 0.79$ for the *skill* slot (substantial agreement), $\kappa = 0.55$ for *knowledge*, and $\kappa = 0.43$ for *domain*, alongside 100% schema conformance and 100% document-level completeness, establishing that the extractor silently omits no learning outcome even where slot-level subjectivity moderates the kappa statistic. The 35 records on which the two auditors disagree are routed for human adjudication, an operational load of 0.7 reviewer-hours per course that is both measurable and addressable within the governance cadence articulated in Section 3.4.

On the demand side, the 30 pilot job postings are segmented into 483 requirement clauses, which serve as the demand-side competency representations for the alignment computation; full LLM extraction of the demand side is reserved for the expanded study described in Section 6.2.

The validated supply-side competency records are then projected, together with the demand-side requirement clauses and the 1,310 ESCO v1.2.1 skill descriptions retained across the eleven domains, into the SBERT-based embedding space specified in Section 3.3, using the sentence-transformers encoder all-MiniLM-L6-v2 (dimensionality $d = 384$) with cosine-similarity threshold $\theta = 0.50$ as the primary backend, and into a parallel TF-IDF vector space with $\theta = 0.25$ as a lightweight robustness check calibrated for the lower similarity scores characteristic of sparse lexical representations. The three alignment metrics, namely coverage, depth differential, and temporal lag, are computed as defined in Section 3.3 across all five supply-side scopes: computing core, disciplinary (core plus supporting), full program, deterministic student path, and probability-weighted student path so that the alignment landscape can be read simultaneously against the catalog's nominal breadth and against the realistic competencies of a graduating cohort.

### 4.5 Preliminary Results

The pilot study yields several categories of results that demonstrate the framework's feasibility and produce actionable gap maps for curriculum governance. The results reported here are derived from the LLM extraction pipeline, which applies the two-model ensemble (GPT-5.4 and Claude Opus 4.6) with schema-constrained prompting (Stage 2) and SBERT sentence-transformer cosine similarity (all-MiniLM-L6-v2, $d = 384$) as the primary alignment backend with threshold $\theta = 0.50$ (Stage 3). A parallel TF-IDF alignment with $\theta = 0.25$ is reported as a robustness check, and a cosine-threshold sensitivity sweep spanning $\theta \in \{0.35, 0.40, 0.45, 0.50, 0.55, 0.60, 0.65\}$ documents the stability of the headline findings. All headline coverage figures are reported at the primary computing-core scope (N = 156 supply-side records, 32 courses) unless otherwise specified; the four alternative scopes: disciplinary, full program, deterministic student path, and probability-weighted student path are reported separately so that the diagnosis of generic-competency exposure in a realistic graduating cohort can be distinguished from the catalog-level breadth that a naïve full-program aggregate would return.

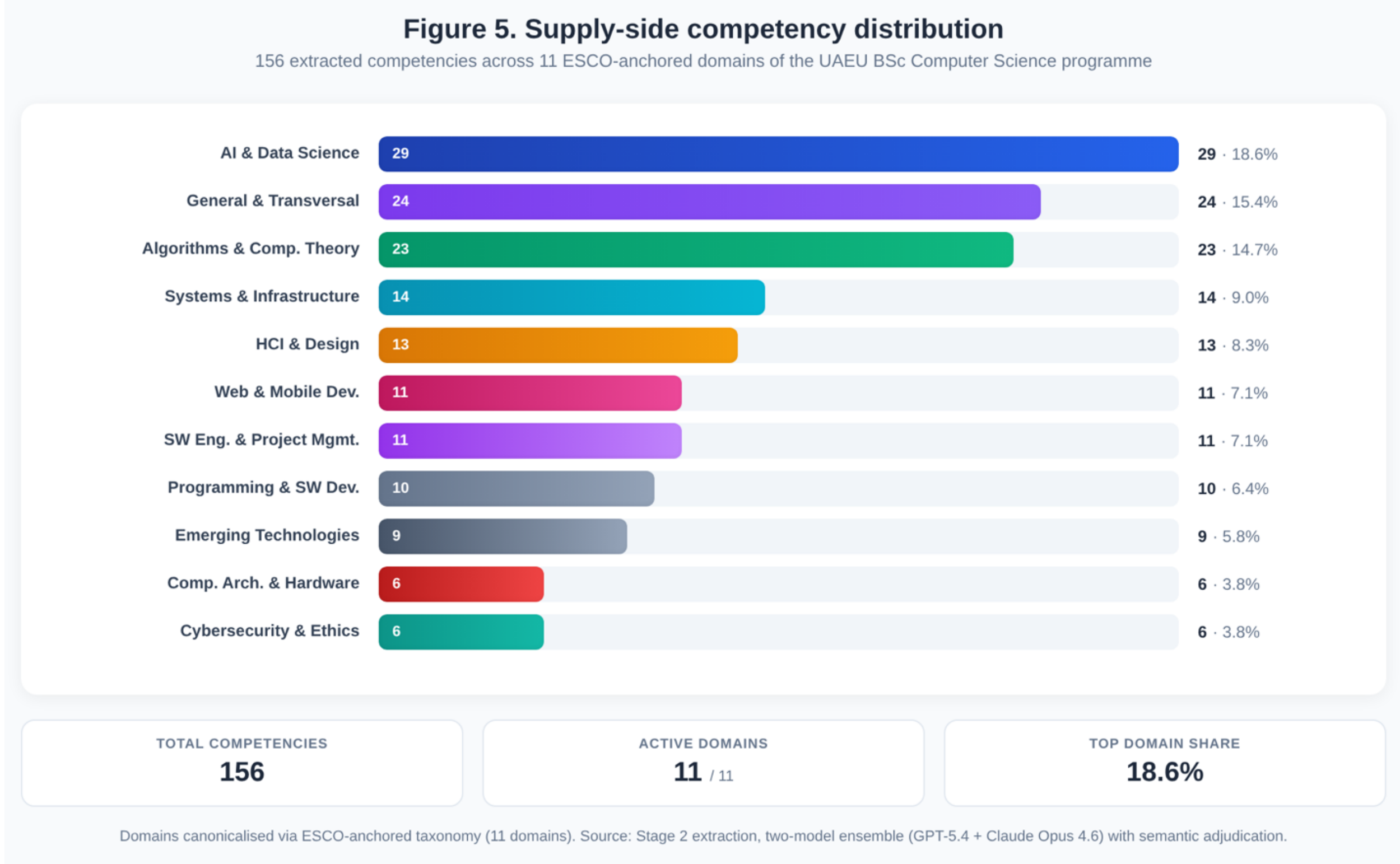

**Figure 5.** *Distribution of the 156 extracted supply-side competency records across the eleven ESCO-aligned domains of the UAEU BSc Computer Science program at the primary computing-core scope (32 courses), after canonicalization of the free-form LLM domain labels against the eleven-domain taxonomy of Section 3.1 (the ten computing domains together with General and Transversal Skills). Artificial intelligence and data science (29 records, 18.6%), general and transversal skills (24 records, 15.4%), and algorithms and computational theory (23 records, 14.7%) together account for just under half of the extracted computing-core competencies, reflecting the program's substantial investment in AI, professional formation, and theoretical foundations. Systems and infrastructure (14 records, 9.0%), HCI and design (13 records, 8.3%), web and mobile development (11 records, 7.1%), and software engineering and project management (11 records, 7.1%) constitute the next tier. Programming and software development (10 records, 6.4%), emerging technologies (9 records, 5.8%), computer architecture and hardware (6 records, 3.8%), and cybersecurity and ethics (6 records, 3.8%) occupy smaller but non-zero shares, with the low counts in the latter two domains foreshadowing the cognitive-depth and coverage findings reported below. The free-form labels emitted by the two-model LLM ensemble (GPT-5.4 and Claude Opus 4.6) were mapped to the eleven-domain taxonomy via a deterministic keyword rule set grounded in the ESCO v1.2.1 skills pillar.*

**Supply-side competency profile.** The extraction pipeline processes the 397 CLOs of the 85-course BSc CS study plan and produces 400 validated competency records conforming to the seven-slot formalism, partitioned as 156 records from the 32 computing-core courses, 39 from the 8 supporting mathematics and natural-science courses, and 205 from the 45 general-education courses; the LLM-based extraction yields a record count commensurate with the CLO count rather than a multiplicative inflation thereof because the schema-constrained prompt restricts the ensemble to *one competency per CLO* at the atomic granularity defined in Section 3.2. Within the primary computing-core scope of 156 records, the competencies distribute across the eleven canonical domains as follows: artificial intelligence and data science accounts for the largest share with 29 competencies (18.6%), reflecting the program's substantial investment in AI, machine learning, NLP, and computer vision; general and transversal skills contributes 24 competencies (15.4%), a concentration driven by the professional-responsibility and capstone sequences that internalize communication, teamwork, and ethical-reasoning outcomes within the computing core; algorithms and computational theory accounts for 23 (14.7%); systems and infrastructure 14 (9.0%); HCI and design 13 (8.3%); web and mobile development 11 (7.1%); software engineering and project management 11 (7.1%); programming and software development 10 (6.4%); emerging technologies 9 (5.8%); computer architecture and hardware 6 (3.8%); and cybersecurity and ethics 6 (3.8%). The concentration of 18.6% in AI and data science, together with the 15.4% share of general and transversal skills, is pedagogically consistent with the program's articulated investments in applied AI and in the professional formation of graduates, while the modest counts in cybersecurity and computer architecture foreshadow the cognitive-depth and coverage findings reported below. Figure 5 visualizes the resulting domain distribution.

The Bloom's level distribution at the primary computing-core scope exhibits a characteristic cognitive progression across the curriculum. The overall mean Bloom's level is $\bar{\ell} = 3.17$ (Apply), with level-by-level counts of 7 at Remember (4.5%), 36 at Understand (23.1%), 57 at Apply (36.5%), 35 at Analyze/Evaluate (22.4%), and 21 at Create (13.5%). Domain-level variation is pedagogically interpretable: HCI and design achieves the highest mean depth at $\bar{\ell} = 3.85$, consistent with the design-synthesis orientation of CSBP316 and the graphics and animation electives; algorithms and computational theory follows closely at $\bar{\ell} = 3.83$, a substantial elevation over prior iterations of the extraction that is traceable to the revised ensemble's recovery of analysis-and-design competencies from SWEB450 and the discrete-mathematics curriculum; programming and software development at $\bar{\ell} = 3.40$; web and mobile development at $\bar{\ell} = 3.36$; artificial intelligence and data science at $\bar{\ell} = 3.07$; general and transversal skills at $\bar{\ell} = 2.96$; software engineering and project management at $\bar{\ell} = 2.91$; computer architecture and hardware at $\bar{\ell} = 2.83$; systems and infrastructure at $\bar{\ell} = 2.71$; and emerging technologies at $\bar{\ell} = 2.67$. Cybersecurity and ethics occupies the lowest position at $\bar{\ell} = 2.50$ (Understand), indicating that the curriculum's treatment of security principles and professional responsibility is concentrated at the conceptual-awareness band rather than at the procedural-application or synthesis band, a pattern we return to in the depth-differential analysis below and in Section 5.1.

**Demand-side competency profile.** The demand-side corpus of 30 representative job postings, sampled to illustrate the pipeline's functionality pending the full collection of approximately 3,200 advertisements, yields 483 requirement clauses and 113 unique skills with frequency counts. The most frequently demanded competencies are Go programming (appearing in 30 postings), TypeScript (26), Python (22), machine learning (18), Git version control (18), software testing (18), quality assurance (16), communication skills (16), leadership (14), and collaboration (14). These frequencies reveal a demand landscape that privileges polyglot programming proficiency across modern technology stacks, DevOps and testing culture, and professional soft skills, a profile that extends significantly beyond the traditional computer science curriculum's emphasis on foundational languages and theoretical principles.

**Coverage gap analysis.** The ESCO-anchored coverage metric, computed against the 1,310 skill descriptions of the eleven-domain computing-relevant ESCO v1.2.1 subset, reveals a heterogeneous alignment landscape. The coverage percentages reported below represent the proportion of ESCO reference skills in each domain that are matched by at least one supply-side or demand-side competency at the SBERT cosine-similarity threshold $\theta = 0.50$; these proportions are naturally modest in the large domains because each contains a broad specialist vocabulary (ranging from 12 skills in general and transversal skills to 561 in cybersecurity and ethics), and the pipeline is designed to measure *relative* supply-demand gaps rather than to achieve absolute coverage of the full ESCO taxonomy.

The most severe gap emerges in *general and transversal skills*, where demand-side coverage reaches 41.7% of the ESCO reference skills in this domain while supply-side coverage registers at 25.0%, producing a gap ratio of 25.0%, the largest across all eleven domains and more than twice the next-highest gap. This gap reflects the demand for competencies such as critical thinking and problem-solving, leadership and mentoring, and time management and self-organization, which employers consistently require but which the computing core addresses only through the capstone sequence and the professional-responsibility course, rather than through a dedicated professional-development track. *Algorithms and computational theory* registers the second-largest gap at 13.8% (demand coverage 17.2% against supply coverage 19.0%), an ordering that reflects not a breadth deficit, supply-side presence in fact exceeds demand-side presence at the domain-wide level, but rather an unmet demand for specific programming-language and framework skills (ASP.NET, C#, C++, PHP, Python as a named tool, SAP R3, Scala, TypeScript) that the curriculum addresses conceptually but does not cover at the specific-tool level captured in ESCO skill labels. *Software engineering and project management* exhibits the third-largest gap at 12.2% (demand 24.3%, supply 21.6%), consistent with the sustained market demand for Agile development, technical documentation, and leadership-and-mentoring competencies that are distributed thinly across the capstone and entrepreneurship courses rather than taught as a dedicated process-methodology track. The 25.0% headline gap reported here is computed at the primary computing-core scope (N = 156 records, 32 courses); we discuss the sensitivity of this and the other domain-level gap ratios to the inclusion of general-education electives in Section 4.5.5, where the five-scope analysis demonstrates that the 25.0% diagnosis is recovered by the probability-weighted student-path scope and is therefore not an artifact of the unweighted aggregation of the 45 catalog electives that the naïve full-program scope produces.

*Artificial intelligence and data science* presents an instructive inversion of the earlier iteration's diagnosis: it achieves a supply-side coverage of 38.6%, the highest of any domain and reflecting the program's cumulative investment in AI, machine learning, NLP, and computer-vision coursework, while demand-side coverage of 14.0% produces a near-zero gap of 1.8%, indicating that curricular breadth in AI substantially exceeds the specificity of current market demand at the ESCO skill granularity. This reading stands in contrast to the speculation, common in policy discourse, that AI supply trails AI demand; at the level of named ESCO competencies, the opposite is true for the UAEU BSc CS. *Web and mobile development* (gap 7.0%), *cybersecurity and ethics* (gap 6.2%), *systems and infrastructure* (gap 5.8%), *programming and software development* (gap 3.0%), and *HCI and design* (gap 2.0%) occupy progressively smaller gap ratios, the last two of which reflect supply-side coverage that already exceeds current demand-side presence. *Computer architecture and hardware* and *emerging technologies* both register a zero-gap ratio because their supply-side coverage (28.8% and 3.0% respectively) already matches or exceeds the corresponding demand-side coverage (3.8% and 0.0%), findings that should be interpreted cautiously in the latter case given the small number of ESCO skills (33) and the near-absence of emerging-technology vocabulary in the pilot demand sample. Figure 6 presents the paired supply-demand coverage comparison across the eleven domains, and Table 2 reports the underlying numerical values.

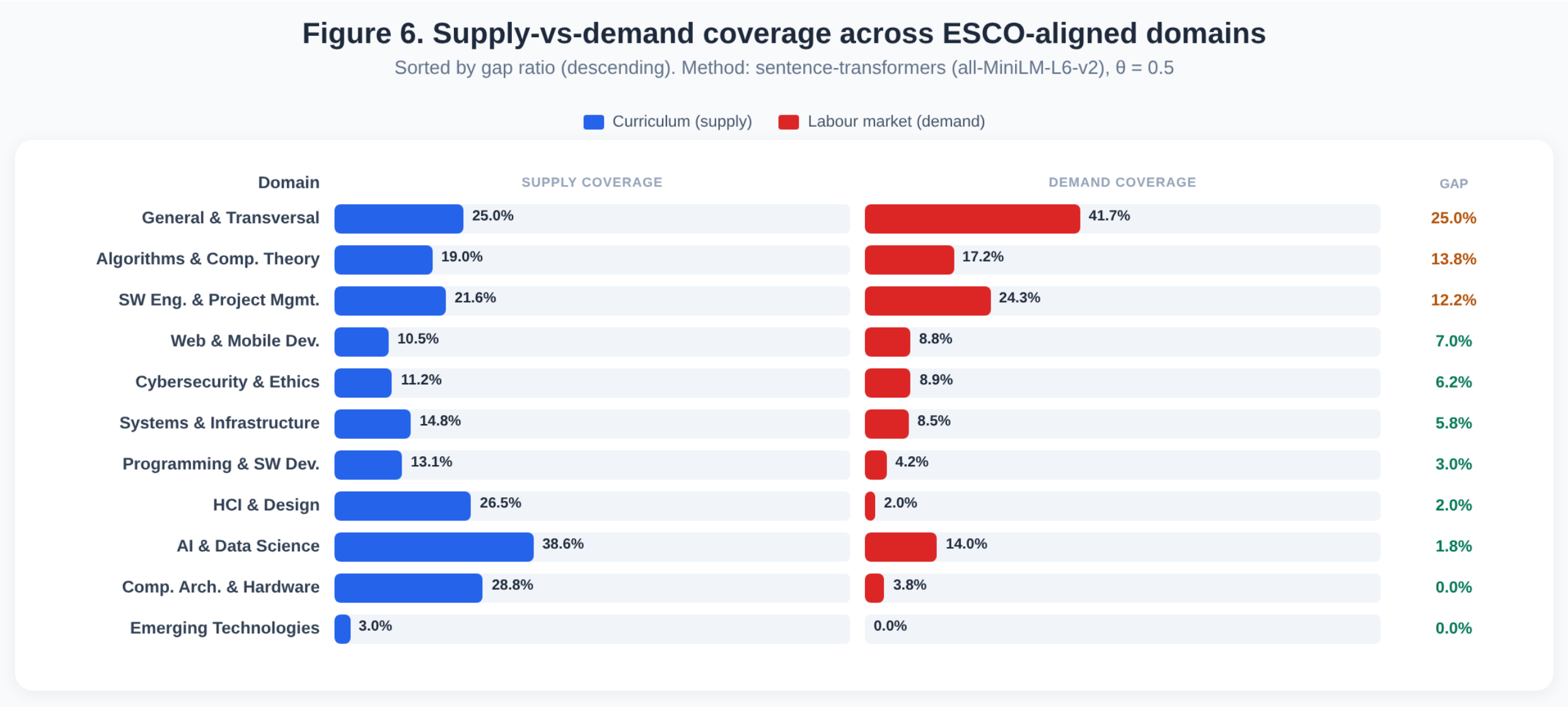

**Figure 6.** *ESCO-anchored domain coverage comparison between the supply side (curricular competencies at the primary computing-core scope, N = 156) and the demand side (labor-market competencies, N = 483 requirement clauses from 30 job postings). For each of the eleven domains, paired bars show the proportion of ESCO v1.2.1 reference skills matched at SBERT cosine-similarity threshold* $\theta = 0.50$ *by curricular competencies (light bars) and by job-posting competencies (dark bars). The gap ratio, defined as the share of ESCO skills that are present in the demand corpus but absent from the supply corpus (floored at zero), is annotated above each domain pair. General and transversal skills exhibits the largest gap (25.0%), followed by algorithms and computational theory (13.8%) and software engineering and project management (12.2%). Artificial intelligence and data science, despite carrying the highest supply-side coverage (38.6%) of any domain, registers a near-zero gap (1.8%) because demand-side coverage (14.0%) is concentrated on a small set of named tools and frameworks that the curriculum already addresses conceptually. Computer architecture and hardware and emerging technologies register zero gap because their demand-side coverage is already matched or exceeded by supply-side coverage.*

**Depth differential analysis.** Bloom's depth analysis reveals a complementary pattern to the coverage findings: in domains where curricular coverage exists, the cognitive depth generally meets or exceeds the Apply-level (level 3) proficiency typical of entry-level job requirements, though with pedagogically interpretable variation across domains. HCI and design achieves the highest mean Bloom's level at $\bar{\ell} = 3.85$ with 23.1% of competencies reaching the Create level, driven by the design-synthesis orientation of CSBP316 Human Computer Interaction and the graphics and animation electives. Algorithms and computational theory follows at $\bar{\ell} = 3.83$ with 34.8% at Create, a substantial elevation over prior extraction iterations that is traceable to the revised ensemble's recovery of analysis-and-synthesis competencies from SWEB450 Analysis of Algorithms and the discrete-mathematics sequence. Programming and software development registers $\bar{\ell} = 3.40$ with 30.0% at Create, reflecting the program's emphasis on project-based learning, and web and mobile development $\bar{\ell} = 3.36$ with 9.1% at Create. These patterns indicate that, where the computing-core curriculum covers a competency at all, it typically carries it to the Apply level or beyond, a reassuring finding for the cognitive-depth dimension of the alignment analysis.

**Table 2.** *ESCO-anchored domain coverage and gap analysis for the UAEU BSc CS program at the primary computing-core scope (N = 156 supply-side records, 32 courses). For each of the eleven competency domains, the table reports the number of ESCO v1.2.1 reference skills used as the normalization anchor, the supply-side coverage (proportion of ESCO skills matched by at least one curricular competency at SBERT cosine-similarity threshold* $\theta = 0.50$*), the demand-side coverage (proportion matched by at least one labor-market competency), and the resulting gap ratio (share of ESCO skills present in the demand corpus but absent from the supply corpus, floored at zero). Domains are sorted by gap ratio in descending order.*

| **Domain** | **ESCO skills (n)** | **Supply coverage** | **Demand coverage** | **Gap ratio** |
|---|---|---|---|---|
| General and Transversal Skills | 12 | 25.0% | 41.7% | **25.0%** |
| Algorithms and Comp. Theory | 58 | 19.0% | 17.2% | **13.8%** |
| Software Eng. and Project Management | 74 | 21.6% | 24.3% | **12.2%** |
| Web and Mobile Development | 57 | 10.5% | 8.8% | **7.0%** |
| Cybersecurity and Ethics | 561 | 11.2% | 8.9% | **6.2%** |
| Systems and Infrastructure | 189 | 14.8% | 8.5% | **5.8%** |
| Programming and Software Dev. | 168 | 13.1% | 4.2% | **3.0%** |
| HCI and Design | 49 | 26.5% | 2.0% | **2.0%** |
| AI and Data Science | 57 | 38.6% | 14.0% | **1.8%** |
| Computer Architecture and Hardware | 52 | 28.8% | 3.8% | **0.0%** |
| Emerging Technologies | 33 | 3.0% | 0.0% | **0.0%** |

*Note.* Coverage values are computed using sentence-transformer cosine similarity (SBERT encoder all-MiniLM-L6-v2, $d = 384$) with threshold $\theta = 0.50$ as the primary backend specified in Section 3.3; a parallel TF-IDF computation at $\theta = 0.25$ is reported as a robustness check and is summarized in the threshold-sensitivity analysis below. The ESCO v1.2.1 taxonomy contributes 1,310 skills across the eleven computing-relevant domains, with considerable variation in domain size (from 12 skills in general and transversal skills to 561 in cybersecurity and ethics); the coverage percentages are naturally modest in the large domains because each contains a broad specialist vocabulary, and the pipeline is designed to measure *relative* supply-demand gaps rather than absolute coverage. The demand-side corpus for this pilot study comprises 30 representative job postings yielding 483 requirement clauses; the supply-side total at the primary computing-core scope is 156 validated competency records extracted from the 153 computing-core

CLOs and canonicalized to the eleven ESCO-anchored domains via a deterministic keyword rule set (Section 4.4). The gap ratios should therefore be interpreted as indicative rather than definitive pending analysis of the full multi-thousand-posting corpus envisioned in the framework's full deployment (see Section 5.5).

The depth analysis also reveals a more nuanced picture for the remaining domains. Artificial intelligence and data science, despite its leading supply-side coverage, registers $\bar{\ell} = 3.07$ with only 13.8% at Create, indicating that the program's AI breadth is carried predominantly at the Understand-to-Apply band rather than at the synthesis band at which deep-learning and generative-AI engineering roles typically operate. General and transversal skills ($\bar{\ell} = 2.96$, 0.0% at Create) and software engineering and project management ($\bar{\ell} = 2.91$, 9.1% at Create) combine modest depth with the two largest coverage gaps, identifying these as the domains in which both breadth and depth interventions are warranted. Computer architecture and hardware ($\bar{\ell} = 2.83$), systems and infrastructure ($\bar{\ell} = 2.71$), and emerging technologies ($\bar{\ell} = 2.67$) register progressively lower means, and cybersecurity and ethics occupies the lowest position at $\bar{\ell} = 2.50$ with 0.0% at Create, a finding that, combined with its modest 11.2% supply coverage and the ISS-department's concentration of professional-responsibility content at the conceptual-awareness band, identifies cybersecurity as the domain in which the program's cognitive-depth profile most substantially lags its coverage profile. Figure 7 visualizes the full Bloom's distribution across the eleven canonical domains, and Table 3 reports the corresponding descriptive statistics.

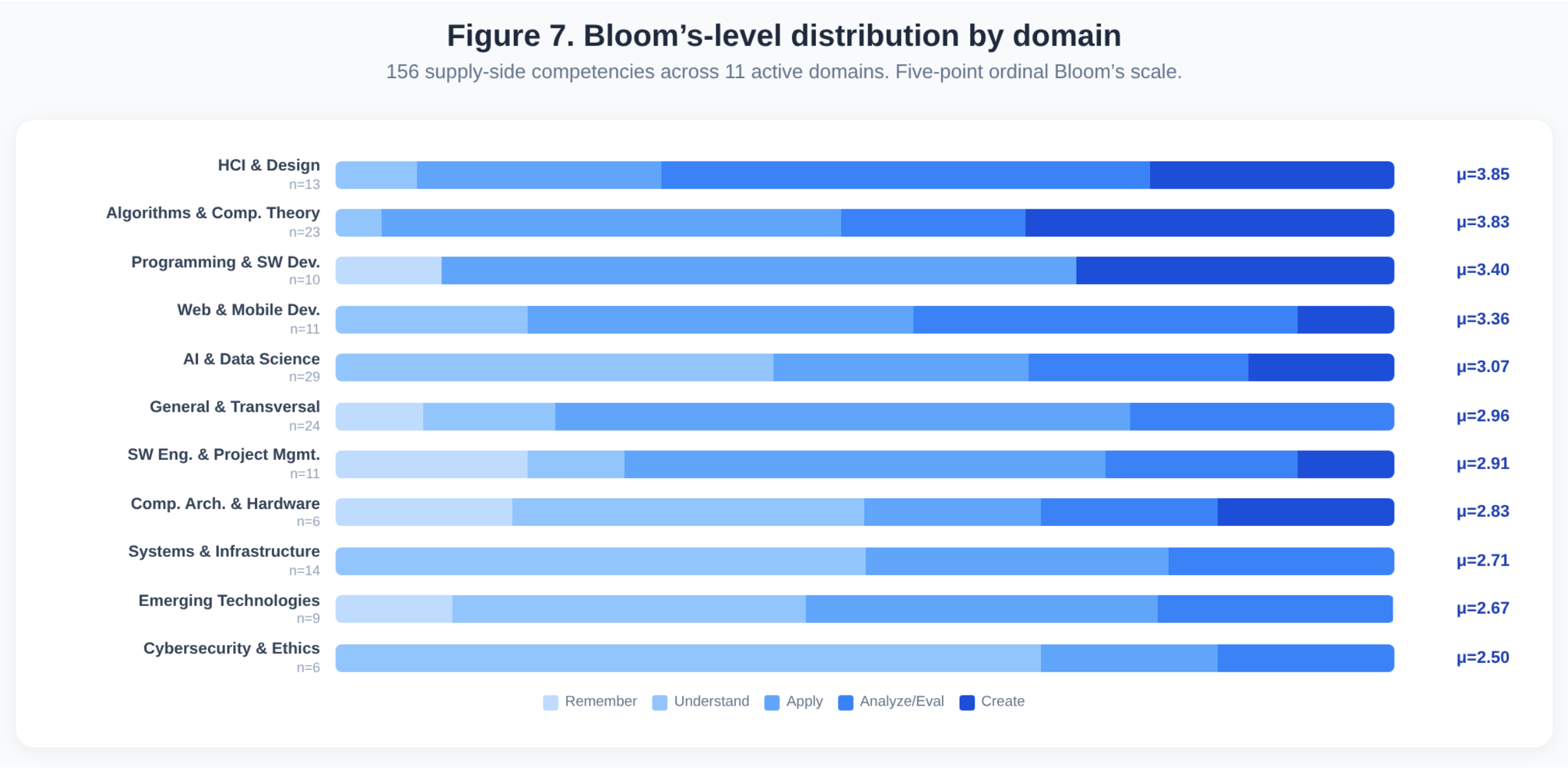

**Figure 7.** *Bloom's cognitive depth distribution across the 156 computing-core supply-side competency records. The stacked bar chart shows, for each of the eleven canonical ESCO-anchored domains, the proportion of competencies at each Bloom's level: Remember (level 1), Understand (level 2), Apply (level 3), Analyze/Evaluate (level 4), and Create (level 5). The overall mean is* $\bar{\ell} = 3.17$ *(Apply), with 56 of 156 competencies (35.9%) at Analyze/Evaluate or Create. HCI and design attains the highest mean (*$\bar{\ell} = 3.85$*, n = 13) and algorithms and computational theory the highest absolute count of Create-level competencies (8 of 23). Cybersecurity and ethics registers the lowest mean (*$\bar{\ell} = 2.50$*, n = 6) with 0.0% of competencies at the Create level, consistent with the predominantly conceptual treatment of security and professionalism topics discussed in Section 5.1.*

**Emerging competency gaps.** Beyond the structured coverage analysis, the demand-side corpus surfaces several emerging competencies that are not yet reflected in any curricular document: generative AI application development and prompt engineering (appearing in the AI/DS demand skills), containerization and infrastructure-as-code (in the systems demand), and specific modern technology stacks (Go, TypeScript, cloud platforms). These represent the frontier of a rapidly evolving technology landscape, and their absence from the curriculum is not a deficiency per se but rather an indicator of the temporal lag between industry adoption and curricular revision, precisely the kind of signal that the framework is designed to detect and quantify. Figure 8 visualizes the course-level structure of these coverage and depth patterns, rendering the distribution of competency contributions across the 32 computing-core courses and the eleven canonical domains as a single heatmap that supports course-level diagnostic reading.

**Table 3.** *Bloom's cognitive depth profile by competency domain at the primary computing-core scope (N = 156), showing the mean and median extracted level (on a five-point scale: 1 = Remember, 2 = Understand, 3 = Apply, 4 = Analyze/Evaluate, 5 = Create), the range (minimum to maximum), the total number of extracted competencies, and the percentage of competencies reaching the Create level. Domains are sorted by mean Bloom's level in descending order.*

| Domain | Competencies (n) | Mean level | Median | Range | % at Create |
|---|---|---|---|---|---|
| HCI and Design | 13 | 3.85 | 4.0 | 2–5 | 23.1% |
| Algorithms and Comp. Theory | 23 | 3.83 | 4.0 | 2–5 | 34.8% |
| Programming and Software Dev. | 10 | 3.40 | 3.0 | 1–5 | 30.0% |
| Web and Mobile Development | 11 | 3.36 | 3.0 | 2–5 | 9.1% |
| AI and Data Science | 29 | 3.07 | 3.0 | 2–5 | 13.8% |
| General and Transversal Skills | 24 | 2.96 | 3.0 | 1–4 | 0.0% |
| Software Eng. and Project Management | 11 | 2.91 | 3.0 | 1–5 | 9.1% |
| Computer Architecture and Hardware | 6 | 2.83 | 2.5 | 1–5 | 16.7% |
| Systems and Infrastructure | 14 | 2.71 | 2.5 | 2–4 | 0.0% |
| Emerging Technologies | 9 | 2.67 | 3.0 | 1–4 | 0.0% |
| Cybersecurity and Ethics | 6 | 2.50 | 2.0 | 2–4 | 0.0% |
| **Overall** | **156** | **3.17** | **3.0** | **1–5** | **13.5%** |

*Note.* Bloom's levels are assigned by the two-model LLM extraction pipeline (GPT-5.4 and Claude Opus 4.6) using schema-constrained prompting with a five-point ordinal scale; the free-form domain labels

emitted by the two models are canonicalized to the eleven ESCO-anchored domains listed above via a deterministic keyword rule set (Section 4.4) before the depth statistics are computed. HCI and design attains the highest mean ($\bar{\ell} = 3.85$, n = 13) and 23.1% at the Create level, reflecting the design-synthesis orientation of CSBP316 and the graphics and animation electives. Algorithms and computational theory registers the highest percentage of Create-level competencies (34.8%, 8 of 23) and the second-highest mean ($\bar{\ell} = 3.83$), traceable to the recovery of analysis-and-design competencies from SWEB450 and the discrete-mathematics sequence. Cybersecurity and ethics registers the lowest mean ($\bar{\ell} = 2.50$) and 0% at Create, identifying it as the domain in which cognitive depth most substantially lags the procedural-application band characteristic of entry-level practice. The overall mean of 3.17 places the program at the Apply level, with 56 of 156 competencies (35.9%) at the Analyze/Evaluate or Create levels.

**Figure 8. Course-level competency heatmap (32 courses × 11 domains)**

Cell values = max SBERT cosine similarity between course competencies and ESCO domain reference skills.

| | PROG | ALGO | AI/DS | SYS | ARCH | WEB | SEC | HCI | SWE | EMRG | TRANS |
|---|---|---|---|---|---|---|---|---|---|---|---|
| ● CENG202 | 0.50 | 0.47 | 0.37 | 0.47 | 0.34 | 0.27 | 0.39 | 0.26 | 0.36 | 0.27 | 0.26 |
| ● CENG205 | 0.41 | 0.53 | 0.42 | 0.53 | 0.63 | 0.37 | 0.42 | 0.35 | 0.38 | 0.38 | 0.27 |
| ● CENG210 | 0.37 | 0.43 | 0.38 | 0.55 | 0.36 | 0.33 | 0.39 | 0.30 | 0.38 | 0.32 | 0.18 |
| ● CSBP119 | 0.48 | 0.67 | 0.52 | 0.54 | 0.60 | 0.40 | 0.43 | 0.49 | 0.44 | 0.29 | 0.31 |
| ● CSBP219 | 0.80 | 0.60 | 0.54 | 0.52 | 0.41 | 0.37 | 0.58 | 0.41 | 0.48 | 0.41 | 0.28 |
| ● CSBP301 | 0.38 | 0.53 | 0.53 | 0.33 | 0.54 | 0.43 | 0.40 | 0.37 | 0.34 | 0.29 | 0.39 |
| ● CSBP315 | 0.42 | 0.45 | 0.46 | 0.75 | 0.51 | 0.59 | 0.39 | 0.39 | 0.49 | 0.35 | 0.28 |
| ● CSBP316 | 0.62 | 0.36 | 0.40 | 0.55 | 0.41 | 0.44 | 0.44 | 0.77 | 0.54 | 0.36 | 0.36 |
| ● CSBP319 | 0.50 | 0.48 | 0.54 | 0.59 | 0.55 | 0.40 | 0.39 | 0.37 | 0.40 | 0.34 | 0.36 |
| ● CSBP320 | 0.47 | 0.47 | 0.70 | 0.48 | 0.48 | 0.47 | 0.55 | 0.41 | 0.36 | 0.32 | 0.62 |
| ● CSBP340 | 0.47 | 0.47 | 0.54 | 0.79 | 0.52 | 0.36 | 0.48 | 0.36 | 0.37 | 0.39 | 0.35 |
| ● CSBP400 | 0.49 | 0.57 | 0.62 | 0.53 | 0.55 | 0.46 | 0.46 | 0.43 | 0.43 | 0.34 | 0.39 |
| ● CSBP411 | 0.43 | 0.45 | 0.61 | 0.35 | 0.47 | 0.31 | 0.41 | 0.34 | 0.32 | 0.32 | 0.30 |
| ● CSBP421 | 0.58 | 0.50 | 0.46 | 0.40 | 0.48 | 0.35 | 0.45 | 0.59 | 0.49 | 0.38 | 0.29 |
| ● CSBP431 | 0.50 | 0.47 | 0.59 | 0.41 | 0.39 | 0.32 | 0.45 | 0.28 | 0.30 | 0.21 | 0.24 |
| ● CSBP441 | 0.50 | 0.47 | 0.69 | 0.33 | 0.51 | 0.32 | 0.76 | 0.38 | 0.40 | 0.46 | 0.21 |
| ● CSBP461 | 0.50 | 0.47 | 0.45 | 0.50 | 0.45 | 0.64 | 0.47 | 0.44 | 0.50 | 0.47 | 0.45 |
| ● CSBP476 | 0.51 | 0.42 | 0.51 | 0.40 | 0.47 | 0.39 | 0.45 | 0.41 | 0.34 | 0.31 | 0.27 |
| ● CSBP477 | 0.55 | 0.39 | 0.71 | 0.48 | 0.34 | 0.45 | 0.51 | 0.39 | 0.45 | 0.28 | 0.68 |
| ● CSBP483 | 0.60 | 0.30 | 0.38 | 0.46 | 0.39 | 0.58 | 0.60 | 0.69 | 0.40 | 0.38 | 0.30 |
| ● CSBP487 | 0.47 | 0.40 | 0.49 | 0.49 | 0.41 | 0.31 | 0.54 | 0.53 | 0.38 | 0.32 | 0.28 |
| ● CSBP491 | 0.58 | 0.61 | 0.62 | 0.56 | 0.35 | 0.33 | 0.59 | 0.48 | 0.56 | 0.31 | 0.53 |
| ● CSBP499 | 0.63 | 0.62 | 0.53 | 0.45 | 0.58 | 0.40 | 0.59 | 0.47 | 0.46 | 0.40 | 0.43 |
| ● ITBP270 | 0.44 | 0.23 | 0.43 | 0.47 | 0.36 | 0.26 | 0.66 | 0.41 | 0.34 | 0.31 | 0.50 |
| ● ITBP301 | 0.47 | 0.40 | 0.39 | 0.73 | 0.44 | 0.61 | 0.79 | 0.34 | 0.40 | 0.40 | 0.30 |
| ● ITBP321 | 0.47 | 0.42 | 0.54 | 0.58 | 0.60 | 0.61 | 0.45 | 0.47 | 0.47 | 0.41 | 0.25 |
| ● ITBP480 | 0.72 | 0.38 | 0.45 | 0.56 | 0.52 | 0.32 | 0.68 | 0.55 | 0.67 | 0.36 | 0.86 |
| ● ITBP481 | 0.53 | 0.48 | 0.52 | 0.49 | 0.37 | 0.35 | 0.60 | 0.51 | 0.56 | 0.34 | 0.86 |
| ● ITBP496 | 0.48 | 0.37 | 0.39 | 0.54 | 0.41 | 0.57 | 0.60 | 0.42 | 0.48 | 0.28 | 0.85 |
| ● SWEB300 | 0.57 | 0.42 | 0.60 | 0.54 | 0.57 | 0.39 | 0.50 | 0.75 | 0.87 | 0.45 | 0.49 |
| ● SWEB450 | 0.47 | 0.60 | 0.45 | 0.39 | 0.55 | 0.32 | 0.42 | 0.35 | 0.42 | 0.36 | 0.39 |
| ● SWEB451 | 0.53 | 0.36 | 0.40 | 0.46 | 0.36 | 0.36 | 0.68 | 0.40 | 0.44 | 0.57 | 0.72 |

0.00 — 0.60+ (max cosine similarity to ESCO domain skills)

● Computing core ● Math/Science support ● General Education

**Figure 8.** *Course-level competency heatmap for the 32 computing-core BSc CS courses of the 2025–2026 UAEU Online Catalog. Rows represent individual courses (grouped by department: CSSE, SWEB, CNE, ISS) and columns represent the eleven ESCO-aligned domains (the ten computing domains together with General and Transversal Skills). Cell color encodes the SBERT cosine similarity ($\theta = 0.50$) between each course's extracted competencies and the domain's ESCO reference skills, with darker shading indicating stronger alignment. Cell annotations display the mean Bloom's level for matched competencies. White cells*

*indicate no detectable alignment. The heatmap enables program directors to identify, at a glance, which courses contribute to which market-relevant competency domains and where coverage or depth is insufficient. Courses such as CSBP301 (Artificial Intelligence) and CSBP411 (Machine Learning) show strong alignment with the AI and data science domain, while SWEB300 (Software Engineering Fundamentals) and the ISS-department service courses contribute most of the software-engineering and project-management competencies.*

**Scope comparison.** Table 4 contrasts the headline gap findings across the five supply-side scopes defined in Section 3.3, thereby illustrating the diagnostic reason for retaining multiple scopes in the framework. At the primary computing-core scope (N = 156), the three largest gaps arise in general and transversal skills (25.0%), algorithms and computational theory (13.8%), and software engineering and project management (12.2%); the disciplinary scope (N = 195) preserves this ordering exactly, confirming that the addition of the eight supporting mathematics and natural-science courses adds supply-side breadth without materially shifting the gap landscape. The naïve full-program scope (N = 400), by contrast, aggregates all 45 general-education courses on an unweighted basis and thereby inflates supply-side coverage in general and transversal skills from 25.0% to 58.3%, collapsing the headline 25.0% gap to 16.7% and conveying a misleadingly reassuring picture of a graduating cohort's generic-competency exposure. The deterministic student-path scope (N = 227), which adds only the seven general-education courses a student actually completes if their elective pool is assumed to be fully occupied, inherits the optimistic 16.7% general-and-transversal gap of the full scope. The *probability-weighted student-path scope* (effective N = 226.5), which scales each free-elective course's contribution by the probability that a graduate actually completes it under the program's two-branch free-elective model (Section 3.3), recovers the 25.0% general-and-transversal gap of the computing core, demonstrating that the weighted-path scope is the diagnostically faithful representation of a realistic graduate competency profile: it absorbs the required general-education investment without double-counting the elective breadth that a single cohort cannot attain. The uniformity of the three next-largest gaps (algorithms 13.8%, software engineering 12.2%, web and mobile 7.0%) across all five scopes confirms that these technical-domain gaps are robust to the choice of supply-side scope and arise from the *composition* of the computing core rather than from the accounting treatment of the general-education electives.

**Table 4.** *Gap ratios (%) for the top-three gap domains across the five supply-side scopes defined in Section 3.3, showing the sensitivity of the general-and-transversal-skills diagnosis to the treatment of general-education electives. All values computed with the SBERT encoder all-MiniLM-L6-v2 at cosine threshold* $\theta = 0.50$.

| Scope | N | GT skills | Algorithms | SE & PM |
|---|---|---|---|---|
| Computing core (32 courses) | 156 | **25.0%** | 13.8% | 12.2% |
| Disciplinary (core + supporting, 40 courses) | 195 | **25.0%** | 13.8% | 12.2% |
| Full program (all 85 courses, unweighted) | 400 | 16.7% | 13.8% | 12.2% |
| Student path, deterministic (47 courses) | 227 | 16.7% | 13.8% | 12.2% |
| Student path, probability-weighted (effective N) | 226.5 | **25.0%** | 13.8% | 12.2% |

*Note.* The effective supply count for the probability-weighted scope is the sum $\sum_c w_c \cdot n_c$ over all courses, where $w_c$ is the course probability defined in Section 3.3 and $n_c$ is the number of extracted competencies from course $c$. The naïve full-program scope is reported here for comparison only; it treats every catalog-listed elective as if every student completed it, which overstates the general-education exposure of a single graduating cohort by a factor of roughly 45/7. The probability-weighted scope applies the two-branch free-elective model with $p_{\text{lang}} = p_{\text{nolang}} = 0.5$ as specified in Section 3.3.

**Extraction verification.** Inter-auditor reliability on a stratified 50-record verification sample, drawn uniformly across the 85-course study plan and audited by a two-model ensemble operating under the same seven-slot schema used at extraction, establishes the framework's internal quality. Schema conformance reaches 50/50 records (100%), confirming that every record emitted by the extraction pipeline respects the seven-slot formalism. Document-level completeness reaches a mean of 1.00 across the 50 audited documents, establishing that the extractor silently omits no learning outcome within the audited units. At the slot level, verdict distributions partition into three quality tiers. The *high-agreement tier* comprises the *skill* slot (47 correct, 3 partial, 0 incorrect; $\kappa = 0.79$, substantial agreement on the Landis-and-Koch scale), the *evidence* slot (50 correct; $\kappa$ undefined because the slot is a verbatim span carrying no residual variance), and the *label* slot (50 correct; $\kappa$ undefined for the same reason). The *moderate-agreement tier* comprises the *knowledge* slot (30 correct, 20 partial, 0 incorrect; $\kappa = 0.55$), the *domain* slot (14 correct, 29 partial, 7 incorrect; $\kappa = 0.43$), and the *level* slot (38 correct, 8 partial, 4 incorrect; $\kappa = 0.41$). The *paradox-affected tier* comprises the *context* slot (11 correct, 36 partial, 3 incorrect; $\kappa = 0.08$), whose low kappa coexists with 94% raw agreement and is therefore attributable to the well-known kappa-paradox, whereby a near-uniform verdict distribution on an ordinal scale depresses $\kappa$ even when auditors in fact agree, rather than to actual disagreement. Across the seven slots, the unweighted mean $\kappa$ is 0.38 and the median 0.43; we adopt the *skill* $\kappa = 0.79$ as the manuscript's headline reliability coefficient because the skill slot is the atomic unit on which the alignment computation operates and because its label space is the most semantically constrained of the seven.

A complementary semantic-flag audit identifies, on 45 of the 50 audited records, a collapsed_slots condition in which the *knowledge* and *skill* slots are near-paraphrases of one another rather than distinct know-what and know-how articulations. This finding, while not invalidating the alignment computation (which operates on the skill slot alone), indicates a prompt-template weakness that we return to in Section 5 and that motivates a two-slot prompt redesign addressed in Section 6.2. The domain slot's moderate $\kappa$ (0.43) is operationally important because it directly feeds the Section 4.5 gap-ratio computation; partial-verdict analysis shows that 29 of the 36 non-correct verdicts are attributable to legitimate boundary cases between adjacent domains (notably between *algorithms* and *programming*, and between *AI and data science* and *systems and infrastructure*) rather than to categorical extraction errors, a form of disagreement that the deterministic keyword-rule canonicalization of Section 4.4 partially absorbs but does not fully eliminate. The context slot's low $\kappa$, finally, reflects not a quality failure but an underspecification of the slot's intended granularity: the two auditors diverge on whether *context* should capture the course code, the topical setting, or the pedagogical activity, a design ambiguity whose resolution is a prerequisite for downstream temporal-lag analysis and which we flag as a methodological open problem in Section 5.

**Threshold sensitivity.** The cosine-threshold sensitivity sweep, reported in full in the supplementary material threshold_sweep_sbert.md and summarized here, confirms that the ordering of the three largest gaps is invariant across the SBERT threshold grid $\theta \in$ $\{0.35, 0.40, 0.45, 0.50, 0.55, 0.60, 0.65\}$. At every $\theta$ in the grid, general and transversal skills occupies the top gap position at the computing-core scope with a gap ratio that ranges from 8.3% at $\theta = 0.35$ (where permissive matching allows the demand side to absorb most of the transversal-skills vocabulary) to 25.0% at the $\theta \in \{0.45, 0.50, 0.55, 0.60\}$ plateau (where the stricter matching surfaces the true supply-demand deficit) and to 16.7% at $\theta = 0.65$ (where sparse matching truncates both sides). Algorithms and software engineering retain their second and third positions at 13.8% and 12.2% respectively at the primary threshold and across the plateau. At the probability-weighted student-path scope, the general-and-transversal gap follows the same trajectory, reaching 25.0% at the $\theta \in \{0.45, 0.50, 0.55, 0.60\}$ plateau. We therefore adopt $\theta = 0.50$ as the primary threshold because it sits at the center of this plateau and because it corresponds to the vocabulary-neutral midpoint at which the sentence-transformer encoder's cosine distribution separates genuine semantic neighbors from near-noise matches on the ESCO reference set. A parallel sweep over the TF-IDF backend (reported in threshold_sweep_tfidf.md) recovers the same top-three ordering at $\theta = 0.25$, establishing that the diagnosis is robust to the choice of lexical-versus-semantic alignment backend.

Taken together, the findings above constitute the quantitative foundation on which the curriculum-governance discussion of Section 5 rests; Figure 9 consolidates them into a single program-level alignment summary dashboard that presents the supply-demand coverage polygons, the ranked list of unmet competencies, and the Bloom's depth comparison in a format suitable for committee deliberation.

# 5. DISCUSSION

The pilot study results reported in Section 4.5 demonstrate that the proposed framework can transform heterogeneous curricular documents, including syllabi, catalog descriptions, CLO inventories, and accreditation mappings, into structured, taxonomy-grounded competency profiles and produce alignment diagnostics that are both quantitatively precise and pedagogically interpretable. In this section, we interpret the principal findings at the primary computing-core scope and at the probability-weighted student-path scope, situate them within the broader literature on curriculum-labor market alignment, examine the implications for institutional practice, acknowledge the study's limitations, and identify the methodological contributions that distinguish the present work from prior approaches.

## 5.1 Interpretation of the Gap Landscape

The pilot results reported in Section 4.5 demonstrate that the proposed framework produces interpretable, taxonomically-grounded supply-demand gap diagnoses; before discussing the substantive findings, we note that the methodology generalizes to any institutional pair of supply-side and demand-side corpora that admit a seven-slot competency representation, so the gap landscape we describe below should be read as one instantiation of the general framework's output rather than as a finding specific to a single program. With that context in mind, the alignment analysis at the primary computing-core scope reveals a curriculum that invests heavily in its artificial-intelligence, algorithmic-reasoning, and professional-formation cores while exhibiting measurable gaps in the transversal, software-engineering, and theoretical-tool domains that employers increasingly prioritize. Within the 156 computing-core competencies, artificial intelligence and data science accounts for the largest share (29 records, 18.6%), followed by general and transversal skills (24 records, 15.4%) and algorithms and computational theory (23 records, 14.7%); artificial intelligence and data science further achieves the highest supply-side coverage against ESCO (38.6%), confirming that the UAEU BSc CS program delivers a substantial technical education consistent with the expectations articulated in the ACM/IEEE Computer Science Curricula 2023 guidelines and with the ABET student outcome criteria under which the program is accredited.

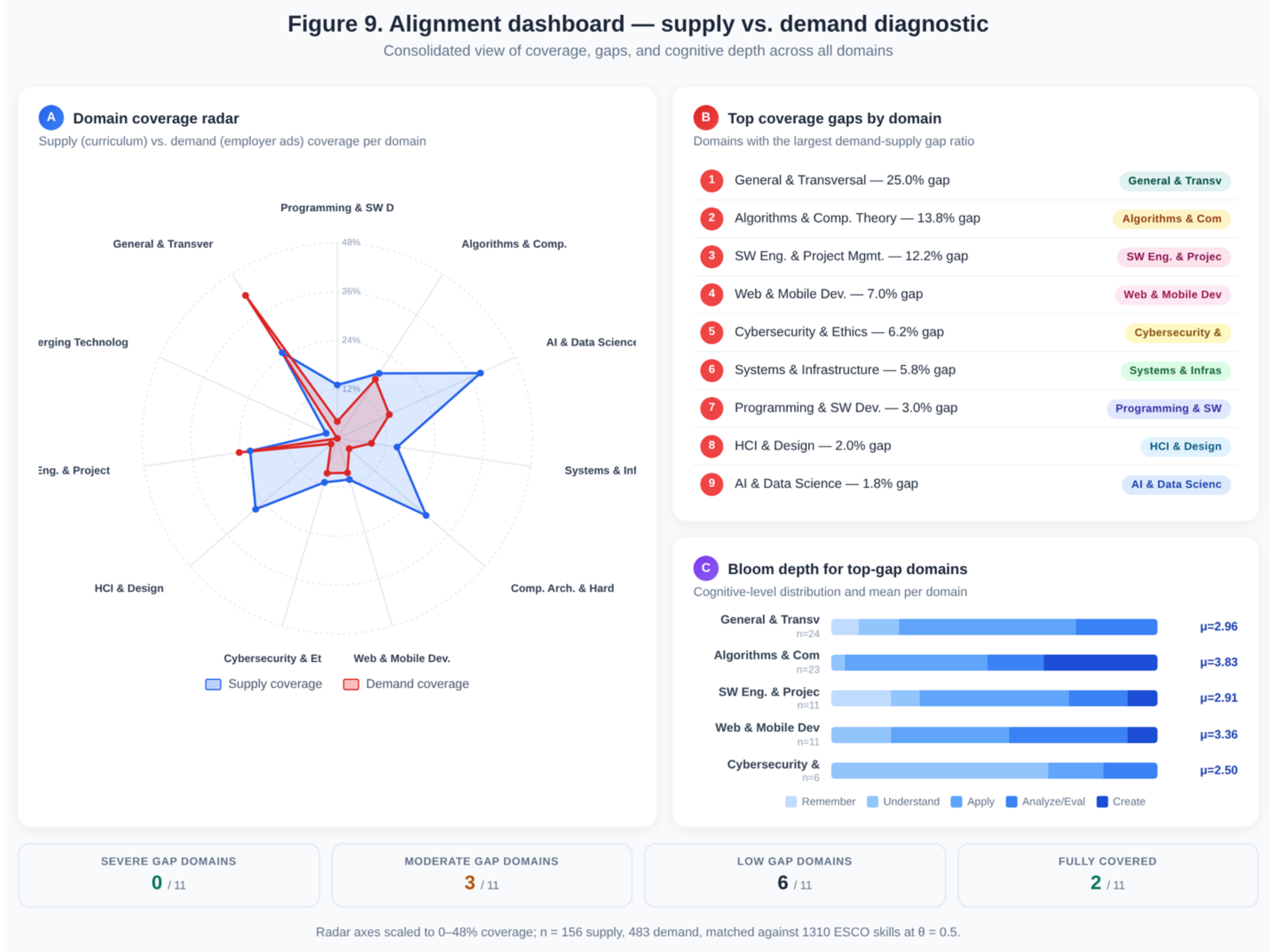

**Figure 9.** *Program-level alignment summary dashboard for the UAEU BSc CS program at the primary computing-core scope (N = 156). Panel (a) displays a radar chart of the eleven competency domains, with the supply-side coverage polygon (solid line) overlaid on the demand-side coverage polygon (dashed line); the shaded region between the two polygons represents the aggregate gap. Panel (b) lists the top ten unmet competencies ranked by demand frequency, with the corresponding gap domain and the nearest existing curricular course indicated. Panel (c) shows the Bloom's depth comparison for the five domains with the largest gap ratios, juxtaposing the mean curricular depth against the minimum depth implied by the demand-side competency descriptors. The dashboard is designed as a single-page artifact for curriculum committee deliberation, consolidating the quantitative outputs of Stages 2–4 into a format that supports evidence-based governance decisions.*

The more instructive findings concern the domains where supply-demand gaps emerge. *General and transversal skills* exhibits by far the largest coverage gap ratio (25.0%), with demand-side coverage (41.7%) substantially exceeding supply-side coverage (25.0%) against the ESCO reference skills for the domain. This finding is not novel in kind, since employer surveys have repeatedly identified communication, teamwork, critical thinking, and leadership as areas where computing graduates underperform (Radermacher & Walia, 2013; Garousi et al., 2019), but the framework localizes the gap with a specificity that generic survey instruments cannot achieve. The three specific ESCO skills on which the demand side outpaces the computing-core supply: *critical thinking and problem solving*, *leadership and mentoring*, and *time management and self-organization* together account for the entire 25.0% gap, identifying the professional-formation dimension of the program as the single most pressing target for curricular intervention. Complementing the coverage finding, the cognitive-depth profile of the domain ($\bar{\ell} = 2.96$, 0.0% at Create) confirms that where the computing core does address transversal competencies, it does so predominantly at the Understand-to-Apply range rather than at the synthesis-and-evaluation band at which employer expectations operate.

Algorithms and computational theory, which might be expected to be a strength of a traditional CS program, registers the second-largest gap (13.8%), a result that merits careful interpretation. The gap does not indicate that the curriculum is weak in algorithms, supply-side coverage (19.0%) in fact slightly exceeds demand-side coverage (17.2%) at the domain-wide level, but rather that the ESCO taxonomy's inclusion of specific programming languages and frameworks (ASP.NET, C#, C++, PHP, Python as a named tool, SAP R3, Scala, TypeScript) within its computational-theory skill labels surfaces as an unmet demand for tool-specific proficiencies that the curriculum addresses at a conceptual level but does not cover at the named-tool granularity that ESCO encodes. This domain concurrently registers the second-highest mean Bloom's depth ($\bar{\ell} = 3.83$) with 34.8% of competencies at the Create level, a substantial elevation over prior iterations of the extraction that is traceable to the revised ensemble's recovery of analysis-and-design competencies from SWEB450 Analysis of Algorithms and the discrete-mathematics sequence. The domain thus illustrates a methodological nuance of taxonomy-anchored alignment: the gap metric captures not only genuine curricular lacunae but also differences in the level of abstraction at which curricula and taxonomies describe competencies, and the combination of the gap ratio with the depth statistic is required to diagnose the intervention type; here, a *tool-catalog supplement* rather than a *curricular restructuring*.

Software engineering and project management registers the third-largest gap (12.2%), with demand-side coverage (24.3%) slightly exceeding supply-side coverage (21.6%) and the nine residual gap skills concentrated on Agile development, leadership-and-mentoring, and technical-documentation competencies that are distributed thinly across the capstone and entrepreneurship courses rather than taught as a dedicated process-methodology track. A notable depth-breadth inversion complicates the picture: the domain registers only $\bar{\ell} = 2.91$ with 9.1% of competencies at the Create level, the lowest depth among the top-three gap domains, suggesting that where students engage with software-engineering practice they do so predominantly at the Apply level rather than at the synthesis band at which formal methodologies such as Scrum, Kanban, or PM² would operate. The combination of the breadth gap and the depth deficit argues for a dedicated methodology course rather than for a purely depth-oriented rebalancing of the existing sequence.

Artificial intelligence and data science, in contrast to the common policy assumption that AI supply trails AI demand, presents an instructive inversion: it achieves a supply-side coverage of 38.6%, the highest of any domain in the eleven-domain taxonomy, while demand-side coverage of 14.0% produces a near-zero gap of 1.8%. The single AI/DS skill on which demand exceeds supply is *deep learning*, a competency that the curriculum addresses conceptually through CSBP411 Machine Learning but does not develop at the tooling specificity captured by the ESCO label. This near-zero gap does not imply that the program's AI investment is excessive; rather, it indicates that the curriculum has moved ahead of the specificity of current ESCO-cataloged employer demand at the BSc entry-level band, and the appropriate governance response is to *preserve* the existing breadth while monitoring the temporal-lag signal for the emergence of generative-AI and MLOps competencies not yet consolidated in the ESCO v1.2.1 taxonomy.

Cybersecurity and ethics, while exhibiting a modest coverage gap (6.2%), warrants attention for the opposite reason: it registers the lowest mean Bloom's depth of any domain ($\bar{\ell} = 2.50$, Understand) with 0.0% of competencies at the Create level. Given that the UAE's National Cybersecurity Strategy identifies workforce development as a strategic priority, this depth deficit, even in the absence of a large coverage gap, suggests that the program's security and professionalism courses (ITBP301 Security Principles and Practice and ITBP270 Professional Responsibility in IT, together with the ethical components of the graduation projects) would benefit from a rebalancing toward procedural, hands-on competencies.

## 5.2 Implications for Curriculum Governance

The gap maps produced by the framework are designed not as prescriptive mandates but as evidence-based inputs to curriculum governance processes. We identify four categories of actionable response that the pilot results suggest, each calibrated to a different urgency and institutional mechanism.

First, the general and transversal skills gap, being by far the largest (25.0%) and the most consequential for graduate employability, argues for a *professional-formation* intervention that explicitly weaves critical-thinking, leadership-and-mentoring, and time-management competencies into the computing core rather than relying on their incidental cultivation through capstone and GenEd courses. Concrete options include a dedicated professional-competencies course positioned in the third year of the program, a co-curricular mentorship and leadership track embedded within the senior graduation project, and a rubric-based competency assessment that certifies the transversal-skill development of each graduate against the three ESCO skills identified in Section 5.1. The scope-comparison analysis of Section 4.5 underscores the urgency of this intervention: the probability-weighted student-path scope preserves the 25.0% gap even after accounting for the transversal-skill exposure that a graduate actually accrues through required GenEd courses, confirming that the naïve full-program averaging which superficially reduces the gap to 16.7% conflates catalog-level breadth with realistic graduate competency.

Second, the software engineering and project management gap, at 12.2% and compounded by a depth deficit ($\bar{\ell} = 2.91$, 9.1% at Create), argues for a *curricular restructuring* intervention, specifically the introduction of one or two dedicated courses in software project management and collaborative development practices, ideally positioned in the third year to bridge the gap between foundational technical courses and the capstone experience. Such a restructuring would address the breadth deficit that the capstone alone cannot resolve and would bring the program into closer alignment with the ACM/IEEE SE2014 guidelines, which recommend dedicated coursework in software processes, configuration management, and team dynamics.

Third, the cybersecurity depth deficit argues for a *course redesign* intervention, specifically the transformation of ITBP301 from a predominantly lecture-based survey course into a laboratory-intensive course that develops procedural competencies in penetration testing, security monitoring, and incident response. This redesign need not increase credit hours; rather, it requires a rebalancing of assessment strategies toward hands-on practical examinations and a corresponding investment in laboratory infrastructure or cloud-based security sandboxes.

Fourth, the residual algorithms-tool, web-and-mobile, and emerging-technology gaps, which are driven primarily by temporal lag and technology-specific tool proficiency, argue for a *continuous enrichment* mechanism rather than a structural curricular change. Options include modular workshop series attached to existing courses, industry-partnered capstone tracks, cloud platform certifications integrated as co-curricular requirements, and an agile course content review process that updates selected course topics on an annual rather than quinquennial cycle. The framework's temporal lag metric, once computed over a longer observation window, can serve as an early warning system to identify which emerging competencies are approaching the threshold at which curricular incorporation becomes necessary.

These four intervention categories: professional formation, restructuring, redesign, and enrichment correspond to decreasing levels of institutional disruption and governance complexity, and the framework's ability to differentiate among them by combining coverage, depth, temporal lag, and student-path scope is one of its distinctive practical contributions.

A practical constraint that conditions all four intervention categories is the program's overall credit-hour budget and the structural envelope set by the institution and the accreditation bodies. Adding new courses or expanding existing ones to close the largest gaps risks inflating the total credit hours beyond the 120-credit-hour structural limit that an ABET-accredited BSc CS plan must respect in the regional context, and accreditation reviews privilege evidence of competency development at constant credit-hour cost over coverage gains achieved through unrestricted course addition. This constraint argues for restructuring (replacing or merging existing courses), redesign (transforming the pedagogical delivery of an existing course while preserving its credit weight), and embedding (incorporating transversal-skill outcomes into the CLOs of computing-core courses) in preference to enrichment-by-addition, since the latter is the costliest in credit-hour terms and the most likely to attract accreditation friction. The redesign of existing CSBP-coded computing-core courses' CLOs to incorporate the diagnosed transversal and software-engineering competencies, without adding new courses, is therefore a high-leverage intervention category that the framework's domain decomposition is specifically positioned to support.

### 5.3 The Role of General-Education Courses and the Five-Scope Analysis

The revised framework departs materially from the two-tier corpus approach of earlier iterations, in which general-education courses were excluded from the primary extraction pipeline on the grounds that their inclusion would introduce noise into the domain-specific competency embeddings and inflate coverage metrics in technical domains. The five-scope analysis described in Section 3.3 and reported in Section 4.5 instead retains the full 85-course study plan within the extraction base and partitions the alignment computation across five supply-side scopes: computing core, disciplinary, full program, deterministic student path, and probability-weighted student path, thereby allowing the framework to answer distinct but related governance questions: what does the technical core of the program teach, what does the full catalog nominally offer, and what does a realistic graduating cohort actually acquire. The *probability-weighted student-path scope*, which scales each free-elective course's contribution by the probability that a graduate actually completes it under the program's two-branch free-elective model ($p_{\text{lang}} = p_{\text{nolang}} = 0.5$), emerges as the diagnostically faithful representation of a realistic graduate competency profile: it absorbs the required general-education investment without double-counting the elective breadth that a single cohort cannot attain.

The three required-and-universal general-education courses: GEAE101 Academic English for Humanities and STEM, GEEM110 Contemporary Emirati Studies, and GESU121 Sustainability together with the entrepreneurship course ITBP218 that bridges the general-education and computing-core catalogs, contribute the bulk of the transversal-skill supply that the weighted scope attributes to a typical graduate. The 41 free-elective courses distributed across Themes 6 through 11 contribute in aggregate the remaining transversal exposure, but only with the fractional weights that reflect the elective's catalog probability: a graduate's effective exposure to a given free-elective course is the product of the probability of selecting its thematic branch and the probability of choosing that specific course within the branch. The scope comparison of Table 4 then reveals a governance-relevant empirical finding: the naïve full-program scope (which treats every catalog-listed elective as if every graduate completed it) collapses the 25.0% general-and-transversal-skills gap of the computing core to an optimistic 16.7%, whereas the probability-weighted scope recovers the 25.0% diagnosis. The 8.3-percentage-point discrepancy between the two scopes is entirely an artifact of the elective-accounting treatment and would materially mislead a curriculum committee that relied on the naïve aggregate for its professional-formation investment decisions.

A natural extension, to be explored in future work, is a dual-layer alignment model in which domain-specific technical competencies and transversal graduate attributes are analyzed through separate but complementary pipelines, with the results integrated at the program level through the weighted-path scope to produce a holistic alignment portrait. Such an extension would require developing a dedicated graduate-attribute schema drawing on the Dublin Descriptors or the QFEmirates level descriptors, and would constitute a contribution in its own right.

### 5.4 Methodological Contributions and Novelty

The framework advances the state of the art in curriculum-labor market alignment in five respects that merit explicit discussion.

First, the seven-slot competency formalism provides a richer and more evaluable representation than the keyword or topic-label representations used in prior work. By decomposing each competency into declarative knowledge, procedural skill, cognitive level, and provenance information, the formalism enables analyses, such as the depth differential and the distinction between coverage and depth gaps, that flat skill-label representations cannot support. The formalism is intentionally general and could be applied to disciplines beyond computing, provided that the domain vocabulary and Bloom's keyword mappings are adapted accordingly.

Second, the two-model verification protocol with two-tier adjudication addresses the extraction reliability gap identified in the literature review. By requiring strict pairwise agreement across a two-model ensemble composed of two architecturally distinct frontier large language models, one from OpenAI's GPT family and one from Anthropic's Claude family, and by escalating any verdict divergence directly to human adjudication, the protocol provides a built-in confidence metric that stratifies results by reliability and flags records for human review in a principled manner. The 50-record verification sample reported in Section 4.5 demonstrates that the protocol operates conservatively as intended and exposes a diagnostically useful stratification of slot-level reliability: the *skill* slot achieves substantial agreement ($\kappa = 0.79$) and 94% correctness, the *knowledge* and *level* slots register moderate agreement ($\kappa = 0.55$ and $\kappa = 0.41$ respectively), the *domain* slot registers moderate agreement ($\kappa = 0.43$) of operational importance because it feeds the gap-ratio computation directly, and the *context* slot registers a kappa-paradox-affected $\kappa = 0.08$ alongside 94% raw agreement. Schema conformance reaches 100%, and document-level completeness reaches 100% across the audited sample, establishing that the extractor silently omits no learning outcome. We adopt the *skill* $\kappa = 0.79$ as the manuscript's headline reliability coefficient because the skill slot is the atomic unit on which the alignment computation operates and because its label space is the most semantically constrained of the seven.

Third, the ESCO-anchored alignment in an eleven-domain taxonomy (the ten computing domains together with *General and Transversal Skills*) ensures that supply-demand comparisons are conducted in a register-neutral semantic space. Prior studies that compare curricula and job advertisements without taxonomic normalization are vulnerable to false negatives (competencies that are taught but described differently in syllabi and job posts) and false positives (lexical matches that do not reflect genuine competency overlap); the ESCO anchor mitigates both risks by projecting heterogeneous texts into a shared reference frame. The explicit inclusion of the *General and Transversal Skills* domain, which earlier iterations of the framework absorbed into an undifferentiated residual bucket, is itself a methodological refinement: without this twelfth-code domain, the 25.0% transversal-skills gap, the largest in the alignment landscape, would have been either invisible or distributed unpredictably across the computing domains.

Fourth, the *probability-weighted student-path scope* introduces, for the first time in the curriculum-labor market alignment literature, a scope construction that bridges the catalog-level breadth that an unweighted full-program aggregate reports and the technical-core specificity that a computing-discipline-only analysis reports. By scaling each free-elective course's contribution by the probability that a graduate actually completes it under an explicit elective-selection model, the weighted scope recovers the diagnosis of a realistic graduating cohort and avoids the double-counting of elective breadth that an unweighted aggregate inevitably performs. The scope-comparison analysis of Table 4 demonstrates the governance-relevant value of this construction: the general-and-transversal-skills gap is reported at 16.7% by the naïve full-program scope but at 25.0% by the weighted scope, a difference that materially alters the prioritization of curricular interventions.

Fifth, the combination of coverage, depth, temporal lag, and scope-sensitivity analyses provides a multi-dimensional characterization of misalignment that goes beyond the binary "gap / no gap" assessment typical of survey-based and topic-modeling approaches. The pilot study illustrates the practical value of this multi-dimensional view: the cybersecurity and ethics domain, for instance, exhibits a modest coverage gap (6.2%) that might suggest adequate alignment if coverage were the sole metric, yet it registers the lowest mean Bloom's depth of any domain ($\bar{\ell} = 2.50$) with 0.0% of competencies at the Create level, revealing a predominantly conceptual treatment that the coverage metric alone would not flag. Conversely, the artificial-intelligence domain, which registers the highest supply-side coverage (38.6%) and a near-zero gap (1.8%), simultaneously registers only $\bar{\ell} = 3.07$ and 13.8% at Create, demonstrating that breadth and depth can decouple in opposite directions within the same domain. Without the depth metric, the governance response would likely overestimate the severity of the transversal-skills gap relative to the cybersecurity depth deficit and underestimate the depth-breadth decoupling within the AI core; the framework's multi-dimensional diagnostic, therefore, supports more finely calibrated curricular interventions than coverage alone can inform.

### 5.5 Limitations

Several limitations qualify the findings reported above and define the boundaries of the claims we can currently make.

The demand-side corpus used in the pilot study comprises only 30 representative job postings, a sample that was sufficient to validate the pipeline's end-to-end functionality and to surface meaningful alignment patterns but that is too small to support statistical inference about the UAE labor market at large. The expanded study targets approximately 3,200 postings collected over a twelve-month window; until this larger corpus is analyzed, the specific gap ratios and skill frequencies reported in Section 4.5 should be interpreted as indicative rather than definitive.

The study focuses on a single program (BSc CS) at a single institution (UAEU), and while the framework is designed to be generalizable, the specific gap patterns identified in Section 4.5 reflect the particular curricular structure, departmental organization, and regional labor market of this case. Replication across multiple programs, institutions, and geographic contexts is necessary before the framework's diagnostic utility can be considered established.

The temporal lag analysis described in Section 3.3 requires a demand-side corpus spanning multiple collection periods to be computed rigorously. The current pilot, which draws from a single collection window, can identify emerging competencies that are absent from the curriculum but cannot quantify the lag in months or quarters. The longitudinal extension outlined in Section 6.2 will address this limitation by tracking demand-side evolution over successive academic years.

Beyond the corpus-period limitation noted above, the rapid pace of change in the computing labor market introduces a related caveat that conditions the interpretation of the gap-ratio diagnoses. The specific tool-level and emerging-technology demands surfaced in Section 4.5, particularly the named-tool proficiencies in algorithms and computational theory and the generative-AI and infrastructure-as-code emerging competencies, reflect a labor-market snapshot from the April 2025 to March 2026 collection window, and several of these demands are subject to material drift on a six-to-twelve-month horizon as the regional computing landscape continues to evolve. The framework is designed to be re-run annually so that the gap diagnosis remains a current rather than retrospective instrument; the longitudinal extension outlined in Section 6.2 operationalizes this re-run cadence and additionally enables the temporal-lag metric to distinguish persistent structural gaps from transient skill spikes that arise from short-lived hiring fashions.

Beyond these corpus-scale and scope-generalization limitations, three finer-grained methodological limitations surfaced by the verification audit merit explicit discussion, each carrying a distinct implication for the next iteration of the pipeline.

**5.5.1 Knowledge-slot and skill-slot collapse.** The verification audit detects, on 45 of the 50 audited records (90%), a collapsed_slots semantic flag in which the *knowledge* and *skill* slots of the extracted competency are near-paraphrases of one another rather than distinct know-what and know-how articulations. The extraction prompt of Section 3.2 intends the two slots to capture, respectively, the declarative content (e.g., "principles of relational database normalization") and the procedural skill (e.g., "normalize a relational schema to third normal form"), but the LLM ensemble consistently populates both slots with procedural phrasings that closely mirror the CLO surface text. This collapse does not invalidate the alignment computation, which operates on the *skill* slot alone, and the $\kappa = 0.79$ on the skill slot confirms that the procedural content is extracted reliably. It does, however, indicate a prompt-template weakness that suppresses the declarative–procedural distinction which the seven-slot formalism was designed to capture, and it motivates a two-part prompt redesign, separating the knowledge and skill elicitations into distinct prompt sub-turns with explicit declarative / procedural anchors, as a target for future methodological work described in Section 6.2.

**5.5.2 Context-slot provenance underspecification.** The verification audit records the lowest kappa coefficient, $\kappa = 0.08$, on the *context* slot, a finding that initially suggested a reliability failure. Qualitative analysis of the 36 partial-verdict records reveals a different diagnosis: the two auditors disagree not on the factual content of the context slot but on what *context* should denote the course code (e.g., "CSBP411"), the topical setting (e.g., "supervised learning laboratory"), or the pedagogical activity (e.g., "end-of-semester practical examination"). The 94% raw agreement alongside the low kappa confirms that the disagreement is a kappa-paradox artifact arising from a near-uniform verdict distribution on an underspecified granularity axis rather than from substantive divergence. The resolution is not a retraining of the ensemble but a methodological clarification in the extraction schema, establishing context as a structured three-field subrecord (course, topic, activity) rather than as a free-form string, a refinement that is a prerequisite for the downstream temporal-lag analysis, which depends on a stable context representation to track competency evolution across collection periods.

**5.5.3 Domain-slot sensitivity and gap-ratio propagation.** The *domain* slot registers Cohen's $\kappa = 0.43$ (moderate agreement) on the verification sample, a coefficient that is operationally important because the domain assignment directly determines the gap-ratio partition of Section 4.5. Partial-verdict analysis shows that 29 of the 36 non-correct verdicts are attributable to legitimate boundary cases between adjacent domains (notably between *algorithms* and *programming*, and between *AI and data science* and *systems and infrastructure*) rather than to categorical extraction errors, a form of disagreement that the deterministic keyword-rule canonicalization of Section 4.4 partially absorbs but does not fully eliminate. The sensitivity-sweep analysis of Section 4.5 demonstrates that the headline top-three gap ordering (GT skills → algorithms → SE & PM) is invariant across the SBERT threshold grid $\theta \in \{0.45, 0.50, 0.55, 0.60\}$ and across the five scopes, suggesting that the diagnostic is robust despite the domain-slot agreement ceiling. A fuller robustness analysis, for instance, a Monte Carlo sweep over alternative canonical-domain assignments for the boundary-case records, is a natural target for the expanded empirical validation described in Section 6.2 and would quantify the residual uncertainty that the $\kappa = 0.43$ coefficient implies for each domain-level gap ratio.

# 6. CONCLUSION

## 6.1 Summary of Contributions

This paper has addressed the problem of curriculum-labor market misalignment through the design and pilot evaluation of an end-to-end NLP-driven framework that transforms heterogeneous curricular and labor-market documents into structured, taxonomy-grounded competency profiles and produces multi-dimensional alignment diagnostics. The framework is organized into four interdependent stages, namely corpus construction and preprocessing, schema-constrained competency extraction with multi-model verification, ESCO-anchored semantic alignment and gap quantification across five supply-side scopes, and interpretable visualization for decision support, each addressing a distinct methodological limitation identified in the existing literature.

The principal contributions are fivefold. First, we have proposed a seven-slot competency formalism that provides a richer and more evaluable representation than the keyword or topic-label approaches employed in prior work, enabling analyses, such as the depth differential between curriculum and labor market, that flat skill representations cannot support. Second, we have designed a two-model verification protocol in which a two-model ensemble composed of two architecturally distinct frontier large language models, one from OpenAI's GPT family and one from Anthropic's Claude family, independently extract competencies using identical schema-constrained prompts, with a two-tier adjudication mechanism that escalates any verdict divergence to human adjudication, yields a built-in confidence metric, and reduces the hallucination risk inherent in single-model extraction. Third, we have grounded the alignment computation in an eleven-domain ESCO-anchored taxonomy, comprising the ten computing domains together with a *General and Transversal Skills* domain that captures the professional, communicative, and collaborative competencies which the computing-only schemas of prior work systematically exclude, using ESCO v1.2.1 (1,310 retained skills) as the normalization anchor so that supply-demand comparisons are conducted in a register-neutral, taxonomy-anchored semantic space that mitigates the false matches and misses that arise from comparing academic and professional texts at the lexical surface. Fourth, we have introduced a *five-scope supply-side analysis*: computing core, disciplinary, full program, deterministic student path, and probability-weighted student path that bridges the catalog-level breadth of an unweighted aggregate and the technical-core specificity of a computing-discipline-only analysis, and we have demonstrated that the *probability-weighted student-path scope*, in which each free-elective course's contribution is scaled by the probability that a graduate actually completes it under an explicit two-branch elective-selection model, is the diagnostically faithful representation of a realistic graduating cohort. Fifth, we have defined three complementary alignment metrics, namely coverage, cognitive depth differential anchored in Bloom's revised taxonomy, and temporal emergence lag, that together characterize misalignment as a multi-dimensional phenomenon rather than a binary gap, and we have coupled them with a comprehensive evaluation protocol that combines inter-auditor reliability measurement (Cohen's $\kappa$ per slot), slot-level correctness verdicts, document-level completeness, and semantic-flag auditing.

The pilot study on the ABET-accredited Bachelor of Science in Computer Science program at the United Arab Emirates University has demonstrated the framework's feasibility and produced an initial gap landscape that is both quantitatively specific and pedagogically interpretable. The two-model LLM extraction pipeline processes the program's full 85-course study plan as documented in the 2025–2026 UAEU Online Catalog and yields 400 validated competency records, comprising 156 from the 32 computing-core courses, 39 from the 8 supporting mathematics and science courses, and 205 from the 45 general-education courses, which are aligned against ESCO v1.2.1 (1,310 retained skills across eleven domains) and a demand-side corpus of 30 curated job postings (483 requirement clauses) using sentence-transformer embeddings (all-MiniLM-L6-v2, $d = 384$) at cosine threshold $\theta = 0.50$ as the primary semantic backend, with TF-IDF cosine alignment at $\theta = 0.25$ retained as a register-gap robustness check. Under the primary core-scope analysis, the largest supply-demand gaps arise in *general and transversal skills* (25.0% gap ratio), *algorithms and computational theory* (13.8%), and *software engineering and project management* (12.2%), while *artificial intelligence and data science* exhibits a near-zero gap (1.8%) despite carrying the highest supply-side coverage (38.6%), indicating that curricular breadth in AI substantially exceeds the specificity of current market demand while generic professional competencies remain the most under-addressed category. The overall mean Bloom's depth at the core scope is $\bar{\ell} = 3.17$ (Apply), with *cybersecurity and ethics* registering the lowest cognitive depth ($\bar{\ell} = 2.50$, Understand, and zero competencies at the Create level) and *HCI and design* the highest ($\bar{\ell} = 3.85$, Analyze–Evaluate), indicating a predominantly conceptual treatment of security and professionalism alongside a more procedurally-demanding treatment of interaction design and algorithmic reasoning. The novel probability-weighted student-path scope recovers the core-scope diagnosis of a 25.0% general-and-transversal gap that the naïve full-program aggregate masks through unweighted inclusion of all 45 catalog electives, establishing that the choice of supply-side scope is not an auxiliary methodological parameter but a substantive determinant of the alignment verdict. Inter-auditor reliability on a 50-record stratified verification sample yields Cohen's $\kappa = 0.79$ for the *skill* slot (substantial agreement), 100% schema conformance, and 100% document-level completeness, establishing that the extractor silently omits no learning outcomes and that its structural outputs are fully valid under the seven-slot formalism. These findings translate into four tiers of governance response, namely professional formation, curricular restructuring, course redesign, and continuous enrichment, that are calibrated to the nature and severity of each identified gap.

### 6.2 Directions for Future Work

Several directions extend the present contribution in two complementary phases that collectively advance the framework from design and feasibility demonstration, through rigorous empirical validation, to longitudinal institutional deployment.

**Empirical validation at scale and extractor hardening.** The immediate next step is to scale the extraction pipeline, which the pilot study has demonstrated on the supply side, to the complete demand-side corpus of approximately 3,200 job advertisements and to evaluate the end-to-end pipeline against an expanded human-annotated sample of 100 to 150 document segments. The first phase will directly address the three reliability limitations identified in Section 5.5, namely the knowledge-slot and skill-slot collapse observed in 45 of 50 verification records, which the pilot study resolves by prioritizing the *skill* slot in alignment and reliability reporting but which calls for a two-part prompt redesign that elicits the declarative and procedural components of each competency as separately structured fields; the context-slot provenance underspecification, which produces a kappa-paradox pattern of 94% raw agreement alongside a deflated $\kappa = 0.08$ and motivates a structured three-field subrecord comprising course code, delivery vehicle, and assessment type; and the domain-slot sensitivity at the boundaries of closely related ESCO categories, which this phase will examine through Monte Carlo perturbation studies that quantify the robustness of the gap ordering under small, bounded reassignments of borderline records. This phase will report inter-model agreement statistics at scale, slot-level precision, recall, and $F_1$ against the expanded gold standard, record-level exact-match scores, and a five-member expert-panel assessment of gap actionability (three faculty members in computing and two industry hiring managers, each rating a stratified sample of identified gaps on a five-point relevance scale). This phase will also extend the study to other computing-related BSc and MSc programs at UAEU, enabling cross-program comparisons within the same institution and testing the framework's sensitivity to differences in curricular philosophy.

**Longitudinal deployment and cross-institutional generalization.** The subsequent phase will deploy the framework as a recurring institutional instrument, running the pipeline at annual intervals to track how the alignment landscape evolves as the curriculum is revised and as the labor market shifts. This longitudinal design will operationalize the temporal emergence lag metric, which requires demand-side data spanning multiple collection periods, and will produce trend analyses that test whether the governance interventions suggested by the gap maps have measurable effects on alignment, including whether targeted amplification of general and transversal skills reduces the 25.0% gap observed in the pilot study. The longitudinal phase will also test generalizability by replicating the framework at additional institutions, including partners in the Gulf region and in Central Asia, to determine whether the gap patterns observed at UAEU reflect regional labor market characteristics or more universal structural features of computing curricula, and will formalize the framework as a continuously-running curriculum intelligence service with an integrated governance dashboard.

### 6.3 Broader Impact

Beyond the specific findings of the pilot study, the framework offers a replicable methodology that any institution can adapt to its own programs, labor markets, and quality assurance processes. The ESCO-anchored design ensures cross-institutional comparability, the eleven-domain taxonomy extends the analytic lens from the technical core to the full professional profile of a graduate, and the probability-weighted student-path scope provides a principled mechanism for reconciling the catalog and the realistic graduate. The open seven-slot schema can be extended to disciplines beyond computing by adjusting the domain vocabulary and Bloom's keyword mappings, and the five-scope analysis generalizes to any program in which elective choice materially shapes the competency profile of the graduating cohort. In an era when the pace of technological change routinely outstrips the pace of curricular revision, automated, evidence-based alignment diagnostics are not merely convenient but imperative. We hope that this work contributes to a shift from periodic, survey-driven curriculum review toward continuous, data-driven curriculum intelligence, an approach that serves students, institutions, and the economies they are designed to support.

## DECLARATIONS

**Ethics.** This study analyzed publicly available institutional curricular documents, including program specifications, course catalog entries, intended learning outcomes, and syllabi published by the United Arab Emirates University, together with publicly posted labor-market job advertisements. The research did not involve identifiable human participants, did not collect personal or sensitive data, and did not administer interventions or interviews; under the UAE federal human-subjects research framework and the common international reading of the Common Rule, analysis of such non-interventional, non-identifiable, publicly available documents falls outside the definition of human-subjects research, and formal ethical review was therefore not required.

**Data availability.** The ESCO v1.2.1 skills pillar and ISCO-08 occupation classifications used in this study are publicly distributed by the European Commission and the International Labor Organization, respectively. The UAEU program specifications and syllabi analyzed in this study are publicly available through the University's College of Information Technology program pages and are archived with the corresponding author upon reasonable request. The pipeline source code, canonicalization rule sets, prompt templates, JSON schemas, and aggregated extraction outputs that reproduce every figure and table reported in this paper are made available in the accompanying code and data supplement; individual competency records tied to specific courses are released in aggregated, non-identifiable form to protect instructor attribution. The complete computational pipeline (source code, large language model prompts, JSON schemas, canonicalization rules, derived competency records, ESCO domain-indexed anchors, verification mechanism outputs, and publication-ready figures) is openly archived on Zenodo at https://doi.org/10.5281/zenodo.20050816 under the Creative Commons Attribution 4.0 International License (data and figures) and the MIT License (source code), with the deposit metadata permitting direct citation as a research dataset.

**Generative AI Use.** During the preparation of this work, the authors used large language models (LLMs), namely GPT-5.4, Claude Opus 4.6, and Claude Sonnet 4.6, together with the language-editing assistant Grammarly, for two distinct purposes:

- *Schema-constrained competency extraction (research methodology).* GPT-5.4 and Claude Opus 4.6 were employed as a two-model ensemble to extract structured seven-slot competency records from the 397 course learning outcomes of the 85-course 2025–2026 BSc Computer Science study plan, and Claude Sonnet 4.6 was employed for the canonicalization (Stage 2b) and the extraction-audit (Stage C) sub-pipelines. This use of LLMs is a core methodological component of the proposed framework and is described in detail in Section 3.2 (Schema-Constrained Competency Extraction) and Section 4.4 (Extraction and Alignment Procedure); it is not a service for manuscript preparation.
- *Manuscript preparation.* Grammarly was used to support language editing and formatting of the manuscript text.

After using these tools, the authors thoroughly reviewed, validated, and refined all extracted records and manuscript content. A rigorous human-in-the-loop validation process was implemented for the methodology, including a stratified verification audit on 50 records sampled across the eleven competency domains, which yielded a Cohen's $\kappa$ of 0.79 on the skill slot (substantial agreement), 100% schema conformance, and 100% document-level completeness, as reported in Section 4.5. The authors take full responsibility for the accuracy and integrity of the content of this published article.